\documentclass[review]{elsarticle}

\usepackage{tikz, tabularx}
\usepackage{multicol, multirow}
\usepackage{amsmath,latexsym,amsbsy,amssymb,amsfonts}
\usepackage{mathtools}
\usepackage{amsthm}
\usepackage{enumerate}
\usepackage{physics}
\usepackage{algorithm}
\usepackage{algpseudocode}
\usepackage{subcaption}
\usepackage{hyperref}
\usepackage{rotating}
\usepackage{array}

\newcommand{\parenthesis}[1]{\left ( #1 \right)}

\def\R{\mathbb{R}}

\def\E{\mathbb{E}}

\renewcommand{\vec}[1]{{\mathchoice
                     {\mbox{\boldmath$\displaystyle{#1}$}}
                     {\mbox{\boldmath$\textstyle{#1}$}}
                     {\mbox{\boldmath$\scriptstyle{#1}$}}
                     {\mbox{\boldmath$\scriptscriptstyle{#1}$}}}}

\journal{European Journal of Control}

\begin{document}

\begin{frontmatter}

\title{Zero-shot Transfer of Reinforcement Learning Control Policies for the Swing-Up and Stabilization of a Cart-Pole System}
\tnotetext[mytitlenote]{This research did not receive any specific grant from funding agencies in the public, commercial, or nonprofit sectors.}

\author[ncsu]{Nikki Xu}

\ead{xxu35@ncsu.edu}

\author[ncsu]{Hien Tran\corref{mycorrespondingauthor}}
\cortext[mycorrespondingauthor]{Corresponding author}
\ead{tran@ncsu.edu}

\address[ncsu]{Department of Mathematics, Box 8205, NC State University, Raleigh, NC 27695}

\begin{abstract}
Reinforcement learning (RL) is a powerful and convenient tool to modernize controller design. 
In this work, we study the zero-shot transfer of RL-based control policies from simulation to hardware for cart-pole swing-up and stabilization. The two policies are trained independently, and the handoff is implemented in Simulink via switching logic. 
We apply a first-order action smoothing filter to prevent hardware damage from high-frequency oscillatory actuation. Pairing this bandwidth-aware filtering with sensitivity-guided domain randomization (DR) and a simple linear curriculum learning (CL) schedule, we obtain a swing-up policy that in all of our experiments injects sufficient energy for handoff into the stabilizer's region of attraction. The stabilization policy rejects disturbances within the tested range, and the swing-up policy can re-engage after larger perturbations and restores the pendulum to the inverted position. 

\end{abstract}


\end{frontmatter}

\section{Introduction}

The cart-pole system is a canonical nonlinear control benchmark: swing-up requires large transient control actions to increase the pendulum energy until it reaches the upright region, where a stabilizer must maintain balance against disturbances. 
Traditionally, the swing-up task can be achieved by position-velocity (PV) control, sliding mode control, energy-based methods, etc. \cite{astrom_swinging_2000, park_swing-up_2009, tum_swing-up_2014, kennedy_real-time_2019}, while the stabilization task can be designed with LQR, PID, and MPC, all of which have been successfully implemented in physical systems \cite{miranda_application_nodate, ozana_design_2012, kennedy_real-time_2016, jezierski_comparison_2017, abeysekera_modelling_2018}.
In recent years, advances in reinforcement learning have gained tremendous success in complex control tasks and sequential decision making, creating new avenues towards controller design \cite{riedmiller_neural_2005, mattner_learn_2012, lillicrap_continuous_2019}. 

The connection between control and reinforcement learning (RL) has long been recognized \cite{sutton_reinforcement_1992, lewis_optimal_2012,polydoros_survey_2017, recht_tour_2019}, but industries are often reluctant to deploy control schemes obtained directly from RL training algorithms. In particular, a virtually-trained control policy can rarely be transferred to a physical system without fine-tuning, largely due to model mismatch, sensor noise, disturbances, etc. This inconsistency in control performance between simulation and reality is sometimes referred to as the ``sim-to-real gap'' \cite{zhao_sim--real_2020, muratore_robot_2022}. There has been a myriad of techniques used by control engineers to reduce the sim-to-real gap \cite{pinto_robust_2017, rusu_sim--real_2018, du_auto-tuned_2021, chattopadhyay_augcal_2024}, including domain randomization (DR) \cite{tobin_domain_2017, peng_sim--real_2018, muratore_domain_2018, muratore_data-efficient_2021, shakerimov_efficient_2023} and curriculum learning (CL) \cite{bengio_curriculum_2009, narvekar_curriculum_2020, marougkas_integrating_2025}, both of which are applied in our controller design. Unfortunately, not enough literature exist that quantifies rigorously the theoretical implications of using these techniques when training for an RL-based controller. Notably though, Chen et al. \cite{chen_understanding_2022} provided some preliminary theoretical investigation into the effect of domain randomization, but certain assumptions on the separability or smoothness must be imposed on a metric over the probability space of Markov decision processes. In our project, we are limited to comparing the sim-to-real transfer using empirical evidence based on experimental performances with different combinations of DR and CL during training, and defer more theoretical discussions to future research.  

On the other hand, promising results of RL-based controller designs have been shown by fine-tuning with real data after applying virtually trained controllers \cite{julian_never_2020, du_auto-tuned_2021, westenbroek_lyapunov_2022, wagenmaker_overcoming_2024}. These fine-tuning approaches often involve applying an exploratory stochastic policy on the real system, and then using the resulting trajectories to help further improve the expected reward. 
In this project, we use reinforcement learning to design two separate controllers for a cart-pole system installed in our lab: one for stabilization and one for swing-up. Simulink is used to deploy the neural network control policies for both tasks as well as to switch between the two control schemes with bumpless transitions.
In particular, our approach does not require any fine-tuning. Instead, we combine a first-order low-pass filter for action-smoothing, a sensitivity-guided domain randomization (DR), and a simple linear curriculum learning (CL) schedule, which yields a swing-up policy that reliably injects sufficient energy for hand-off into the stabilizer's region of attraction; while the stabilizer is designed previously with a physically-realistic, high-fidelity model that matches our lab equipment.

The organization of this paper is as follows: Section 2 covers some background materials for reinforcement learning and its application to controls, as well as some techniques for improving sim-to-real transfer of RL policies; Section 3 contains initial training including the stabilization control and some experimental results, as well as a first RL-based swing-up control which failed during deployment;  Section 4 reports our sensitivity analysis aimed at identifying suitable model parameters for domain randomization in order to improve the sim-to-real transfer; Section 5 contains a case study to compare the practical effects of domain randomization and curriculum learning; finally, the Appendix contains some additional figures for the switched swing-up and stabilization cart-pole system as implemented in the lab. 

\section{Background}
In this section, we introduce the environment dynamics for training RL policies, some basic background on reinforcement learning for continuous control tasks, and the training algorithms that we use to design controllers.
\subsection{Environment Dynamics}

\begin{table}[h!]
    \centering
    \begin{tabular}{c|c|c}
    \hline
        Variable Description & symbol & Unit \\
    \hline
    \hline
        Position of Cart & $x$ & m\\
        Velocity of Cart & $\dot{x}$ & m/s \\
        Acceleration of Cart & $\ddot{x}$ & m/s$^2$ \\
        Angle of Pendulum & $\alpha$ & rad \\
        Angular Velocity of Pendulum & $\dot{\alpha}$ & rad/s \\
        Angular Acceleration of Pendulum & $\ddot{\alpha}$ & rad/s$^2$ \\
        Voltage Applied to  Motor & $V_m$ & V \\
    \hline
    \hline
    \end{tabular}
    \caption{List of Variables}
    \label{tab:TableOfVars}
\end{table}
\begin{table}[h!]
    \centering
    \begin{tabular}{c|c|c}
    \hline
        Parameter Description & symbol & Value \\
    \hline
    \hline
        Mass of Cart & $m_c$ & 0.94 kg\\
        Mass of Pendulum & $m_p$ & 0.23 kg\\
        Length of Pendulum & $l_p$ & 0.3302 m\\
        Gravitational Acceleration & $g$ & 9.8 m/s$^2$\\
        Radius of Motor Pinion & $r_{mp}$ & $6.35 \cdot 10^{-3}$ m \\
        Motor Armature Resistance  & $R_m$ & 2.6 $\Omega$ \\
        Equivalent Viscous Damping at Hinge & $B_p$ & 0.0024 N-m-s/rad \\
        Equivalent Viscous Damping on Motor Pinion & $B_c$ & 5.4 N-m-s/rad \\
        Gear Ratio & $K_g$ & 3.71 \\
        Electromotive Force (EMF) Constant & $K_m$ & $7.67 \cdot 10^{-3}$ V-s/rad \\
        Motor Torque Constant & $K_t$ & $7.67 \cdot 10^{-3}$ N-m/A \\
        Rotational Moment of Inertia of Motor’s Shaft & $J_m$ & $3.90 \cdot 10^{-7}~$kg-$m^2$ \\
        Pendulum’s Moment of Inertia at Hinge & $J_p$ & $3.344 \cdot 10^{-2}$ ~ kg-m$^2$ \\
    \hline
    \hline
    \end{tabular}
    \caption{List of Parameters}
    \label{tab:TableOfParam}
\end{table}

\begin{figure}
    \centering
    \resizebox{0.6\textwidth}{!}{  %
        \begin{tikzpicture}
    \shade[bottom color=gray!50, top color=white] (-1,0.25) rectangle (16, -0.1); %
    \draw[thick] (-1,0.25) -- (16,0.25); %

    \newcommand{\cartwidth}{1.7} %
    \newcommand{\cartheight}{0.8} %
    \newcommand{\tracky}{0.3}   %
    \newcommand{\pendulumlength}{4} %
    \newcommand{\wheelradius}{0.15} %

    \newcommand{\drawcart}[4]{
        \draw[fill=gray!50] (#1, #2) rectangle ++(\cartwidth, \cartheight); 
        \draw[fill=black] (#1+0.35, #2) circle (\wheelradius); %
        \draw[fill=black] (#1+\cartwidth-0.35, #2) circle (\wheelradius); %
        \draw[ultra thick] (#1+\cartwidth/2, #2+\cartheight) -- ++(#3:#4); %
    }

    \drawcart{0}{\tracky}{270}{\pendulumlength};

    \drawcart{4}{\tracky}{225}{\pendulumlength};

    \drawcart{8}{\tracky}{45}{\pendulumlength};

    \drawcart{12}{\tracky}{90}{\pendulumlength};

    \draw[->, thick] (15, -2) -- ++(1, 0) node[right] {\(x\)};
    \draw[->, thick] (15, -2) -- ++(0, 1) node[above] {\(y\)};
\end{tikzpicture}
    }
    \caption{Sketch of Environment}
    \label{fig:Sketch}
\end{figure}
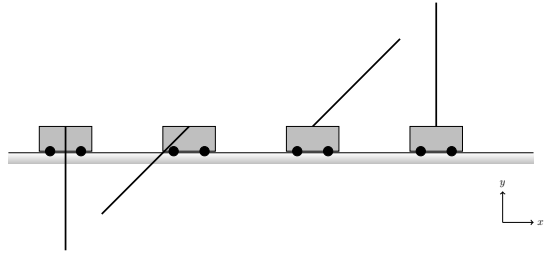

\begin{figure}
    \centering
    \includegraphics[width=0.5\linewidth]{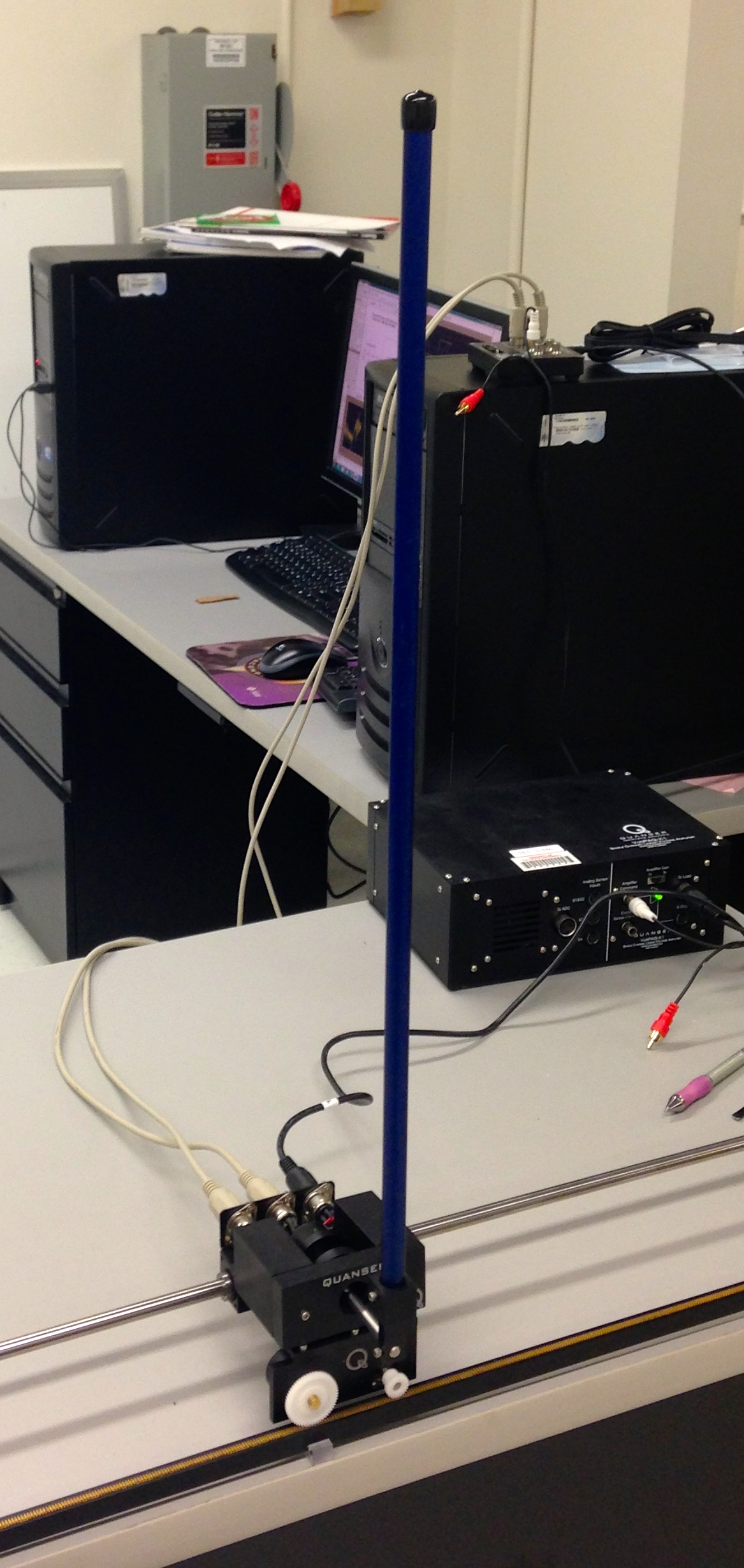}
    \caption{Lab Photo}
    \label{fig:lab_photo}
\end{figure}

The cart-pole system consists of a cart of mass $m_c$ that can move horizontally along a finite-length track and a pendulum of mass $m_p$ and length $2l_p$ that is attached to the cart with a hinge. 
The pendulum is free to swing in the vertical plane. The system is actuated by applying a voltage $V_m$ to a DC motor that drives the cart. 
The positive direction is defined as the cart moving to the right and the pendulum rotating counterclockwise. The cart position, $x$, is zero in the middle of the track, and the pendulum angle, $\alpha$, is zero in the upright position. 
The swing-up task requires moving the pendulum from the stable downward equilibrium ($\alpha = -\pi$) to the unstable upright equilibrium ($\alpha = 0$). 
The stabilization task requires maintaining the pendulum in an upright position against small disturbances.
The equation of motion that governs the dynamics of this system is derived from a Lagrangian mechanics first principle approach, which can be found in~\cite{kennedy_real-time_2016,xu_control_2025}. 
The variables are listed in Table~\ref{tab:TableOfVars}, and the parameters along with the physical meaning of each symbol used in the model are listed in Table~\ref{tab:TableOfParam}. A schematic of the system is shown in Figure~\ref{fig:Sketch}, and a photo of the lab installation is shown in Figure~\ref{fig:lab_photo}. 

The system can be derived using the Euler-Lagrange equation, and after incorporating the actuator dynamics, the second-order ODEs that govern the dynamics of the cart-pole system in the lab can be modeled as follows:
\begin{subequations}
\label{eq:lab_dynamics}
    \begin{multline}
        \ddot{x} = \frac{1}{d} \parenthesis{
            -\frac{4}{3} m_p l_p^2 (B_{eq} \dot{x} + m_p l_p \cos(\alpha) \sin(\alpha) \dot{\alpha}^2) + m_p l_p \cos(\alpha) (-B_p)\dot{\alpha}
        } \\
        +
        \frac{1}{d} \parenthesis{
            \frac{4}{3} m_p l_p^2 C_r V_m + m_p^2 l_p^2 \cos(\alpha) \sin(\alpha) g
        },
    \end{multline}
    \begin{multline}
        \ddot{\alpha} = \frac{1}{d} \parenthesis{
            m_p l_p \cos(\alpha) \parenthesis{
                -B_{eq}\dot{x} + m_p l_p \sin(\alpha) \dot{\alpha}^2 }
            + \parenthesis{m_c+m_p+J_{eq}}(-B_p \dot{\alpha})
        } \\
        +
        \frac{1}{d} \parenthesis{
            m_p l_p \cos(\alpha) C_r V_m + \parenthesis{m_c + m_p + J_{eq}} m_p l_p \sin(\alpha) g
        },
    \end{multline}
\end{subequations}
where $ d = \parenthesis{\frac{4}{3} m_p l_p^2 ( m_c + m_p + J_{eq} ) - ( m_p l_p \cos(\alpha) )^2 }$ is the determinant of the mass matrix in the second-order term. $B_{eq}$ is the equivalent damping force in the cart in the presence of voltage supply, and we define $C_r$, $J_{eq}$, and $K_{eq}$ as constants to shorten the notation
\[
J_{eq} = J_m \parenthesis{\frac{K_g}{r_{mp}}}^2, 
K_{eq} = \frac{K_g^2 K_t K_m}{R_m r_{mp}^2}, 
B_{eq} = B_c + K_{eq}, \text{ and } 
C_r = \frac{K_g K_t}{R_m r_{mp}}.
\]
In both the swing-up and stabilization tasks, the controls are the voltage supply, $V_m$, which is designed to be a feedback control in the form of deep ReLU neural networks whose inputs are state variables. 

\subsection{Reinforcement Learning and Control System}

Our goal is to design controllers for ODE systems with reinforcement learning. Traditionally, RL problems are formulated as Markov decision processes characterized by a tuple $(S,A,P,R,\gamma)$, where: 
\begin{itemize}
    \item $S$ is the set of states (equivalent to the state variables $x$);
    \item $A$ is the set of actions (equivalent to the control variables $u$);
    \item a transition probability function, $P(s, a, s')$, denotes the probability that action $a$ in state $s$ will lead to state $s'$ in the next time step and corresponds to the right-hand side of the dynamical system to be controlled; 
    \item a reward function, $R(s, a, s')$, denotes the reward received after transitioning from state $s$ to $s'$ with action $a$ at each step $t$;
    \item and in some cases, including infinite horizons, an additional discount factor $\gamma \in (0,1)$ is included.
\end{itemize}

A policy $\pi : S \to A$ maps states to actions if it is a deterministic policy, or maps states to a distribution to actions if it is a stochastic one. We denote by $\pi(a \mid s)$ the probability of selecting action $a \in A$ given state $s \in S$. The sequence of states (and in some literature, including actions) is called a trajectory, which we denote by $\tau$, and the subscripts on state and action denote the time step in the trajectory. We write $\tau \sim \pi$ to describe a trajectory $\tau$ where the action at every step is sampled according to a stochastic policy $\pi$, that is, $a_t \sim \pi(\cdot \mid s_t)$; and for a deterministic policy, the action is taken according to the policy exactly at every time step, i.e., $a_t = \pi(s_t)$. This serves as feedback control and, together with the dynamics of the environment, determines the next state $s_{t+1} = f(s_t, a_t)$.

Then, the goal of a finite horizon optimal control problem can be reformulated into an RL problem aimed at maximizing the cumulative reward over a finite trajectory. That is, the objective function becomes the following.
\begin{equation}
    \label{eq:exp_reward}
    \underset{\pi}{\textbf{maximize } } 
    \underset{\tau \sim \pi}{\E} \left[ \sum_{t=0}^{T-1} \gamma^t R(s_{t+1} \mid s_t, a_t) \right] 
\end{equation}
if the policy is stochastic, or 
\begin{equation}
    \label{eq:determ_reward}
    \underset{\pi}{\textbf{maximize } } \sum_{t=0}^{T-1} \gamma^t R(s_{t+1} \mid s_t, a_t), 
\end{equation}
if the policy is deterministic. 

In this project, our swing-up controller is designed with a deterministic policy with stochasticity built into the underlying environment, whereas our stabilization controller is designed with a stochastic policy with the underlying environment being fixed and has no uncertainty. 

\subsection{Training Techniques}
\label{sec:techniques_overview}
We introduce some training techniques beyond the basic RL algorithms that aim at improving sim-to-real transfer performance. 

\subsubsection{Low-pass filter}
Low-pass filters have been used in signal processing and control systems for decades \cite{hayt_engineering_2012}, and can be interpreted as a type of action smoothing in the RL context. Notably, the nature of RL in optimal control theory is closely related to dynamic programming, and the maximization of cumulative rewards for a discretized control system is prone to producing bang-bang control. Thus, when the discretization step size is small--which usually improves the simulation numerical accuracy--it can also induce artificially high-frequency controller input as a result. In our training environment, we apply a simple first-order low-pass filter with a fixed cut-off frequency to the neural network control output signal. This filter is critical for protecting our lab equipment from damage, which will be discussed in the next section. 
\subsubsection{Domain randomization}
Domain randomization~\cite{tobin_domain_2017} is a simple idea, since model parameters are often associated with uncertainties, one can sample near the nominal values of those parameters during training and thus reduce the sensitivity of the policies learned on those fixed parameters. 
This technique has been widely adopted and has been shown to be quite successful in transferring a policy from simulation to real world implementation~\cite{muratore_domain_2018, ramos_bayessim_2019,muratore_assessing_2021}. 
Some of the successfully transferred policies are ``fine-tuned'' with real world data \cite{westenbroek_lyapunov_2022}, and Zhong et al. \cite{zhong_survey_2025} recently wrote a comprehensive summary for these types of data augmentation used for RL training. 
Others have achieved great robotics manipulation with zero-shot transfer \cite{li_reinforcement_2024, zakka_mujoco_2025}, which is training a controller in a virtual environment and directly applying the resulting policy to the real world without fine-tuning. 
This will also be our approach in this project. There has also been theoretical work that shows the effectiveness of domain randomization in improving the robustness of a policy \cite{zhong_pac_2019, chen_understanding_2022}. 
To the best of our knowledge, all known results on the bound of sim-to-real gap with domain randomization rely on finiteness of at least the action space. 
However, in most robotics control schemes, including ours, the control action can be continuous, which makes the analysis significantly more difficult. 
In this work, we choose to randomize the model parameters by uniformly sampling in an interval near their nominal values. 
\subsubsection{Curriculum Learning}
Curriculum learning was first introduced by Bengio et al.~\cite{bengio_curriculum_2009}, and subsequently popularized in the RL community for a wide variety of applications~\cite{narvekar_curriculum_2020}. 
The basic idea of curriculum learning is to gradually increase the diversity of training examples so that the more difficult tasks or environments are introduced later in the training. 
This training method enjoys the benefit of stabilizing initial training due to reduced initial stochasticity, as well as potentially leading to better final performance. 
In this work, we implement a simple linear curriculum schedule that gradually increases the uncertainty margins of the randomized parameters over some fixed amount of training episodes. 
We also experimented with different combinations of domain randomization and curriculum learning to see how they affect the final performance of the trained controllers, but a more rigorous theoretical investigation lies outside the scope of this paper. 

\section{Initial Application of RL-based Controllers}
Without any of the sim-to-real transfer techniques listed above in Section~\ref{sec:techniques_overview}, we were able to obtain a deployable stabilization controller from a classic RL algorithm based on sufficiently accurate modeling of the training environment. However, the swing-up task could not be achieved with the same approach due to hardware failures. As we observe in Section~\ref{sec:stab}, using a high-fidelity model of the cart-pole system combined with the Runge–Kutta numerical integration method (RK4), the trained controller from on a simple RL algorithm can be transferred directly to the physical hardware and maintains the pendulum upright under moderate disturbances. By contrast, the swing-up controller resulting from a similar training setting failed on hardware due to the bang-bang control behavior damaging the plastic wheel on the cart. 

\subsection{Stabilization Controller Design}
\label{sec:stab}
In our training environment for stabilization control using reinforcement learning, the transition probability function is the first-order discretized ODE system based on Equations~\ref{eq:lab_dynamics}, with a fixed step size of 0.01 second. 
The state at each time step $t$ is $s_t = (x, \dot{x}, \alpha, \dot{\alpha})$, corresponding to the state variables of a typical state-space representation of the cart-pole system. 
The reward function is the indicator function over the set of states such that $\abs{x}<0.2$ meter and $\abs{\alpha}<0.2$ radian, representing the cart staying close to the center of the track and the pendulum being close to the upright position, respectively. 
The action at each time step, $a_t$, is the control input to the dynamical system, i.e., the voltage supply to the cart's motor which can vary continuously between $\pm 10$ volts as specified by the manufacturer. 
The maximum step of each episode is set to 2000 steps, corresponding to a 20-second simulation trajectory. The initial condition of each training episode is sampled uniformly within $\pm0.08$ for all four state variables, which mimics the initial condition of the dynamical system in the lab where the velocities are low, pendulum is nearly upright, and the cart is close to the center of the rail. The control input then interacts with the discretized dynamical system to produce an observation of the next time step, and a reward is evaluated based on the new observation. The discount factor is set at 0.99. Finally, training is considered to be finished when the controller successfully maintains the cart on track and balances the pendulum for 20 seconds 95\% of the time over 5000 episodes. 

In the Python gymnasium suite~\cite{towers_gymnasium_2023}, a custom environment is created based on the information outlined above. We use the vanilla policy gradient algorithm,  REINFORCE~\cite{williams_simple_1992},  
to train for a stabilization control policy parameterized by a fully-connected feedforward deep ReLU neural network. The neural network takes the state variable at each time $t$ as input, and outputs the mean and standard deviation of a normal distribution, $\mathcal{N}(\mu, \sigma)$, from which the control signal $a_t$ is sampled. This neural network can therefore be interpreted as a feedback control law in classical control theory literature. An optimization algorithm, NAdam~\cite{dozat_incorporating_2016}, is then used to update the weights and biases of the neural network to maximize the expected discounted cumulative reward. 

\begin{figure}[htbp]
     \centering
     \begin{subfigure}[b]{0.49\textwidth}
         \centering
         \includegraphics[width=\textwidth]{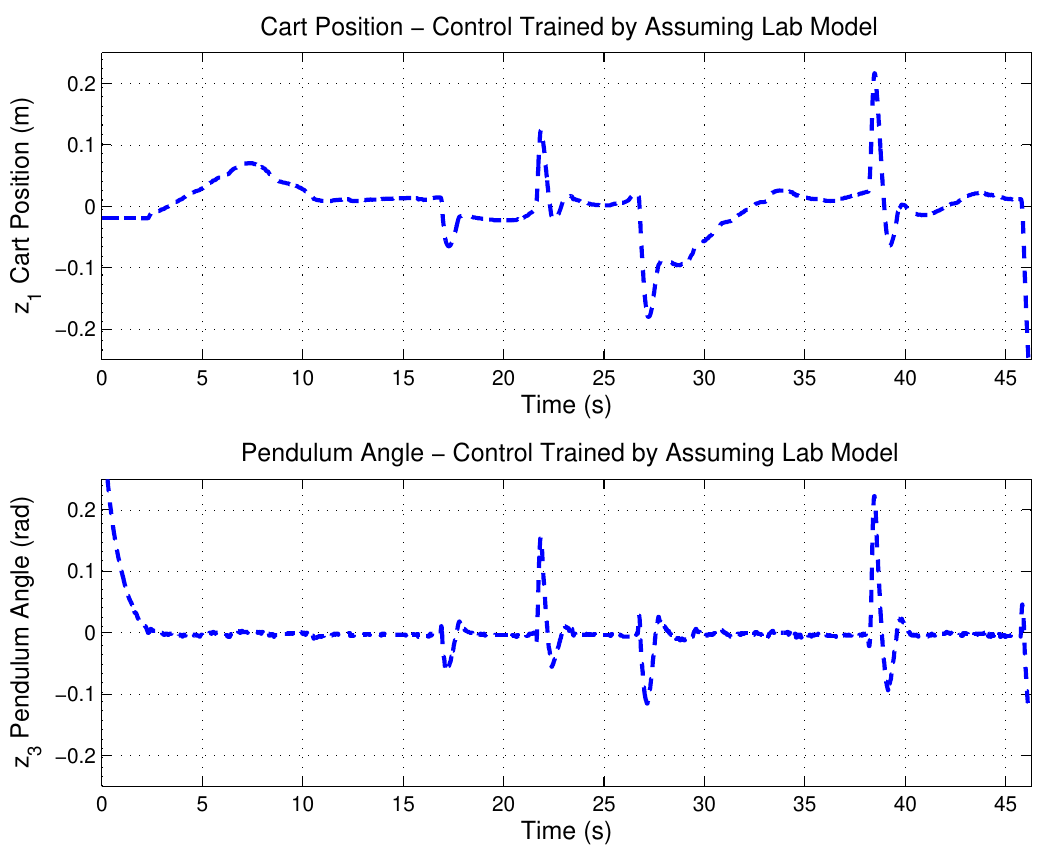}
         \caption{run \#1}
         \label{fig:NonLinControl_run1}
     \end{subfigure}
     \hspace{0.1\textwidth}
     \begin{subfigure}[b]{0.49\textwidth}
         \centering
         \includegraphics[width=\textwidth]{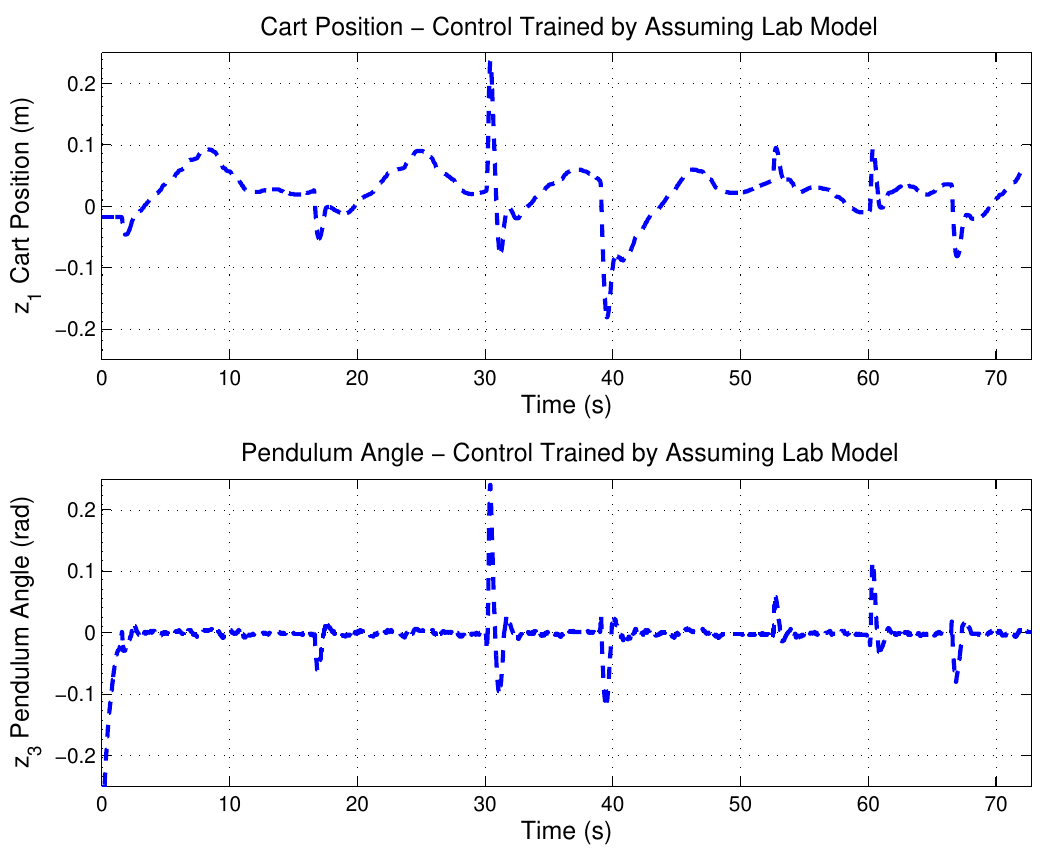}
         \caption{run \#2}
         \label{fig:NonLinControl_run2}
     \end{subfigure}
     \begin{subfigure}[b]{0.49\textwidth}
         \centering
         \includegraphics[width=\textwidth]{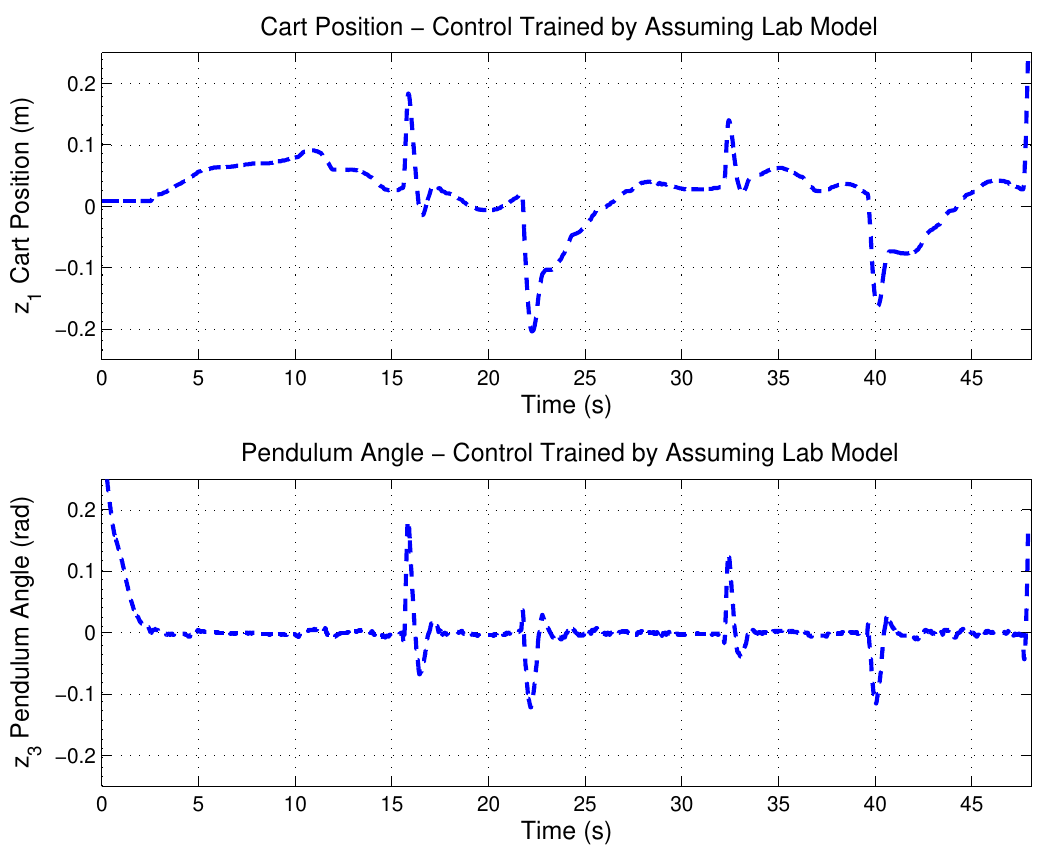}
         \caption{run \#3}
         \label{fig:NonLinControl_run3}
     \end{subfigure}
     \hspace{0.1\textwidth}
     \begin{subfigure}[b]{0.49\textwidth}
         \centering
         \includegraphics[width=\textwidth]{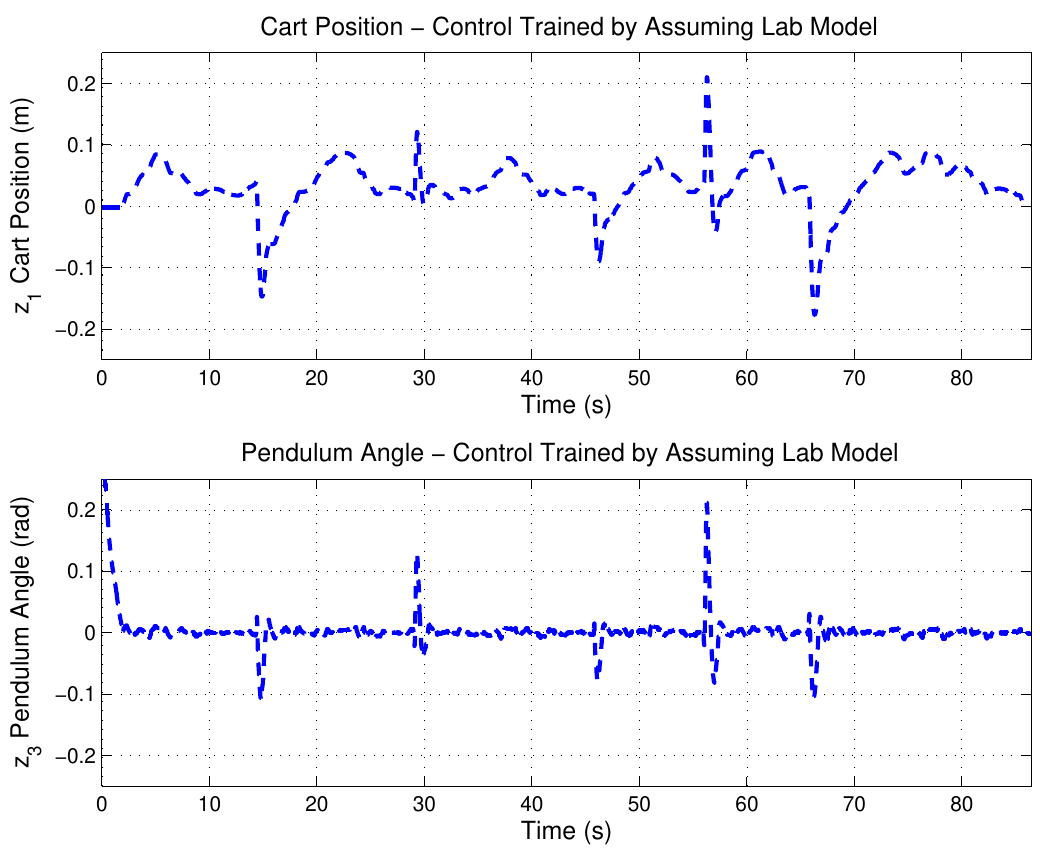}
         \caption{run \#4}
         \label{fig:NonLinControl_run4}
     \end{subfigure}
     \begin{subfigure}[b]{0.49\textwidth}
         \centering
         \includegraphics[width=\textwidth]{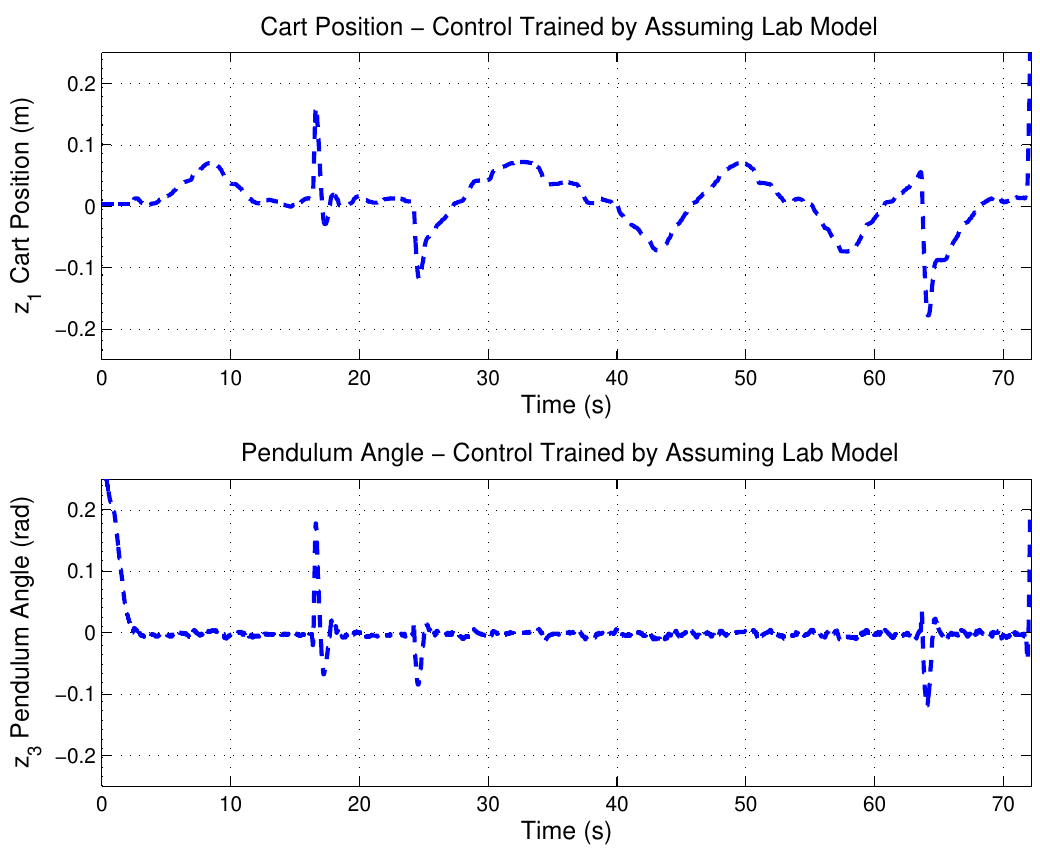}
         \caption{run \#5}
         \label{fig:NonLinControl_run5}
     \end{subfigure}
        \caption{Control learned with lab model tested in lab. Top plot of each subfigure is the position of the cart, and bottom plot of each subfigure is the pendulum angle. The spikes correspond to manually tapping the pendulum to test robustness. }
        \label{fig:FullModel_Experiment}
\end{figure}

Once the training is done, we load the matrices of weights and biases of the resulting neural network into Matlab and Simulink to serve as the feedback control law for balancing the cart-pole system. The implemented control signal is deterministic, i.e., without the standard deviation output from the neural network. Therefore, the feedback control law is a piecewise affine function of the state, $x_t$. We record five trajectories of some exemplary lab experiments in Figure~\ref{fig:FullModel_Experiment}. Each subplot contains the measured cart position on the top and the measured pendulum position on the bottom. The spikes in pendulum positions in each subplot correspond to the pendulum being tapped manually. In all five experiments, we observe that the controller was able to quickly stabilize the pendulum when it was gently tapped up to approximately 14 degrees from the upright position. 

\subsection{Swing-up Controller Design}
\label{sec:swing}

Once the stabilization controller is obtained, we move on to designing a separate controller for the swing-up task. This time we use the Twin Delayed Deep Deterministic Policy Gradient (TD3) algorithm \cite{fujimoto_addressing_2018} from the Python stable-baseline3 library~\cite{raffin_stable-baselines3_2021}, and again using a customized cart-pole environment to simulate the discretized ODE system of our lab model in Equation~\ref{eq:lab_dynamics}. The state observations are $(x, \dot{x}, \sin(\alpha), \cos(\alpha), \dot{\alpha}) \in \R^5$, and the control action, $V_m$, lies in a continuous interval that represents the voltage applied to the motor. TD3 algorithm uses a deterministic policy with additive noise for exploration. We arbitrarily chose two deep ReLU neural networks for the actor and the critic of the algorithm. The reward at each time step is (almost) quadratic with normalized weights as follows:
\[
R(x,u) = -0.1 \parenthesis{\frac{x}{0.25}}^2 - 0.4  \parenthesis{\frac{\dot{x}}{0.25}}^2 - 10  \parenthesis{\frac{\alpha}{\pi} }^2 - 0.1  \parenthesis{\frac{\dot{\alpha}}{\pi} }^2 - 0.01 u^2 - 100 B, 
\]
where the indicator function $B$ equals 1 if the cart position is more than 0.23 meter away from center (runs off track), and 0 otherwise. The numerical integrator was chosen to be a fixed-step RK4, with a time increment of 0.01 second, consistent with our Simulink hardware-in-loop integrator. 

\begin{figure}[htbp]
    \begin{subfigure}[t]{0.5\textwidth}
        \includegraphics[scale=0.28]{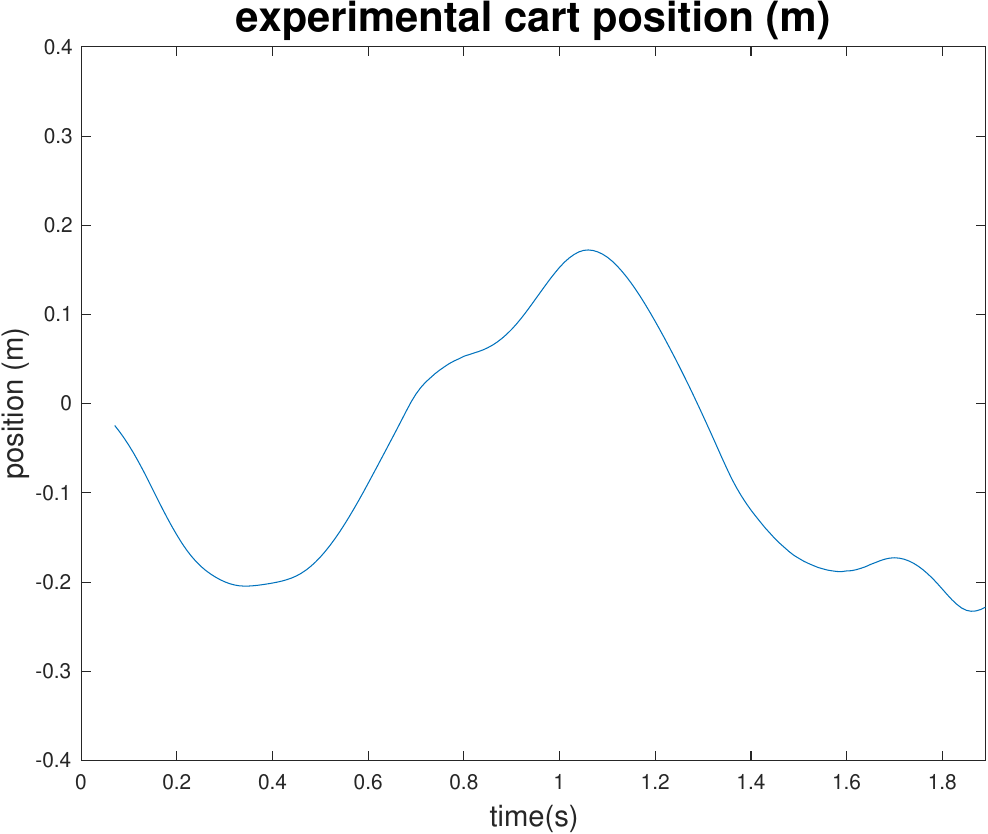}
        \label{fig:swing_old_x}    
    \end{subfigure}%
    \hfill
    \begin{subfigure}[t]{0.5\textwidth}
        \includegraphics[scale=0.28]{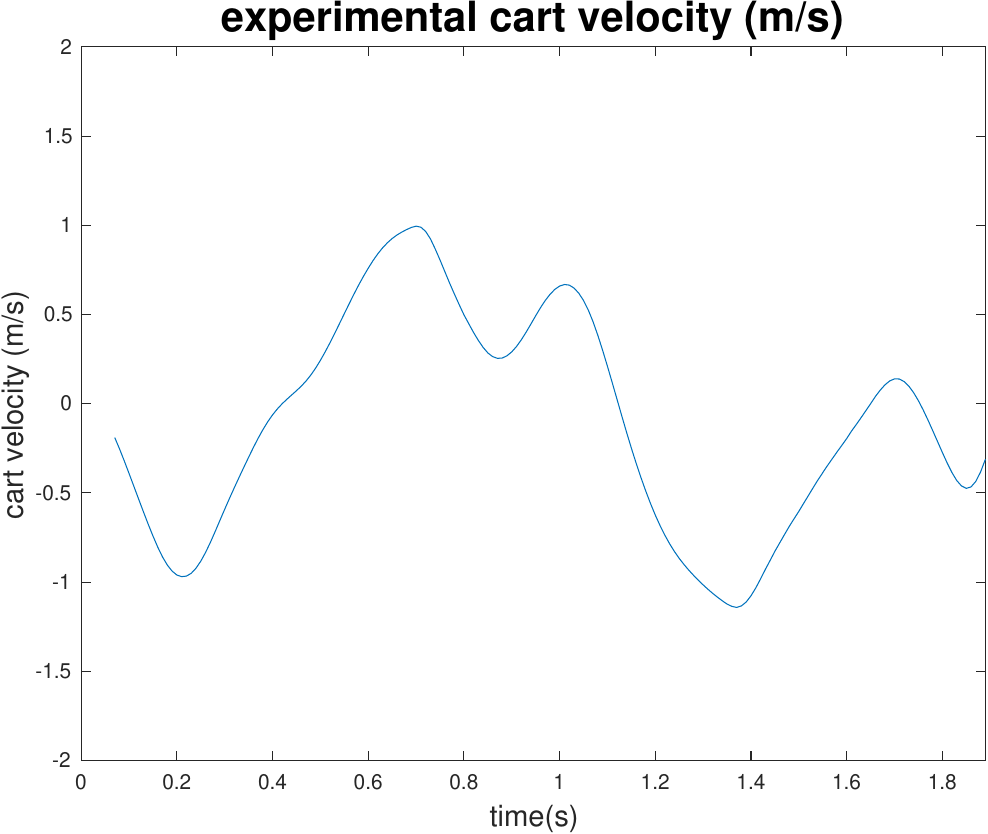}
        \label{fig:swing_old_xdot}    
    \end{subfigure}
    \begin{subfigure}[t]{0.5\textwidth}
        \includegraphics[scale=0.28]{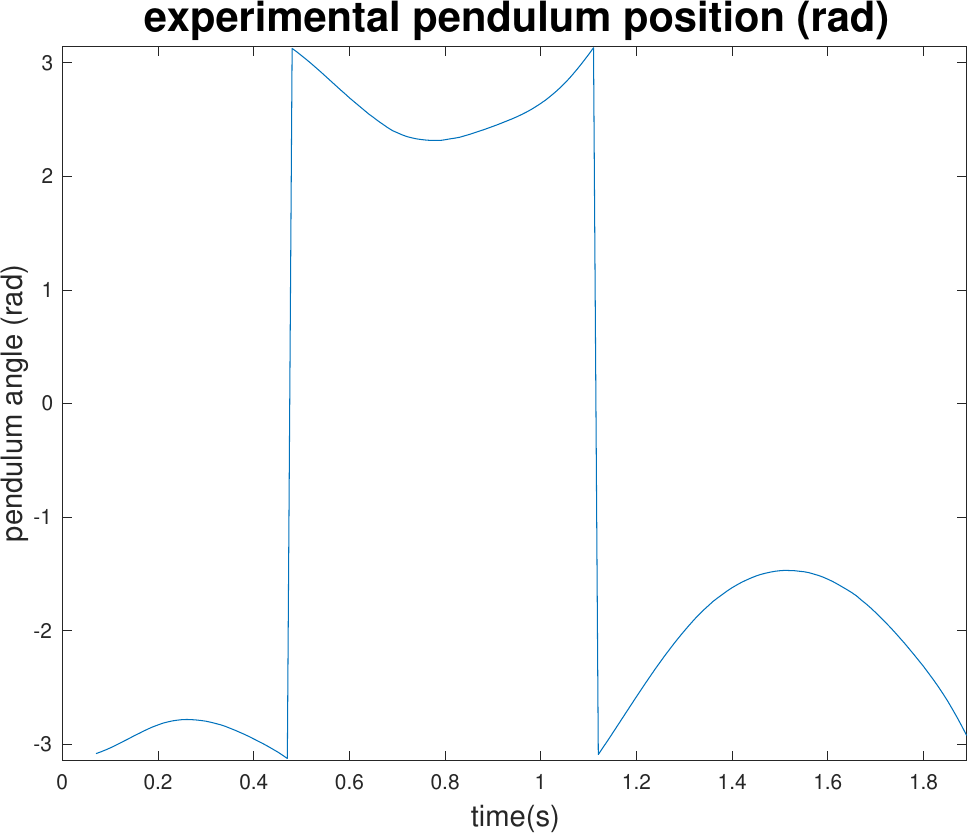}
        \label{fig:swing_old_alpha}    
    \end{subfigure}%
    \hfill
    \begin{subfigure}[t]{0.5\textwidth}
        \includegraphics[scale=0.28]{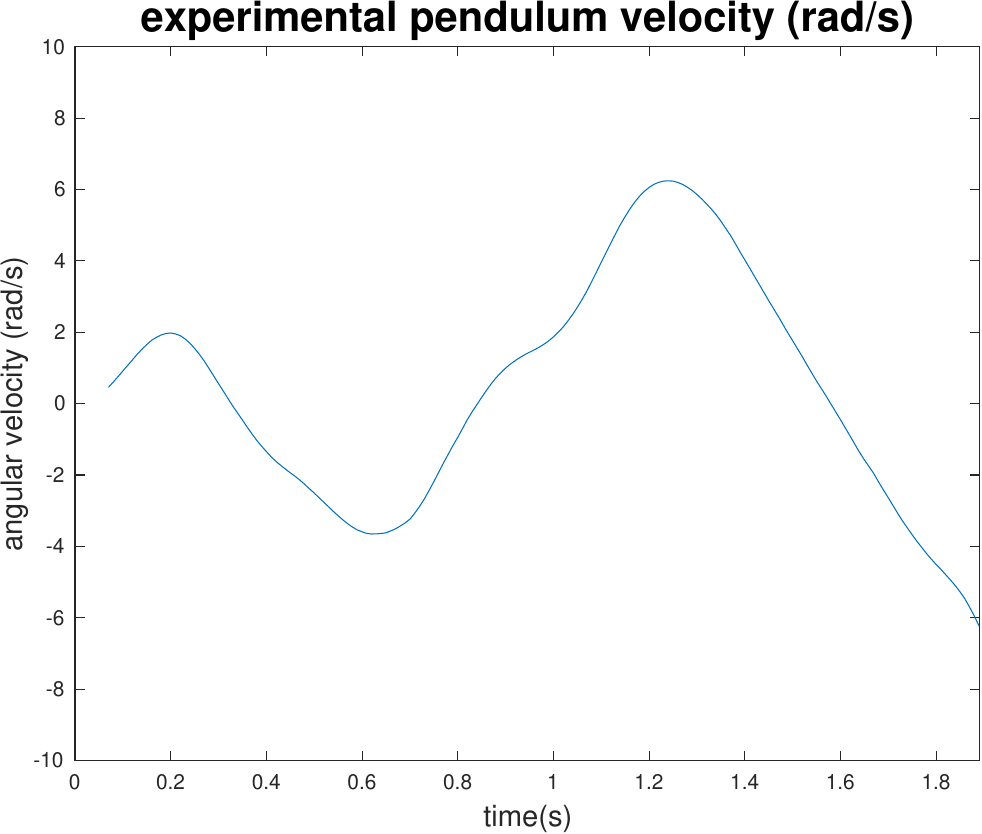}
        \label{fig:swing_old_alphadot}    
    \end{subfigure}
    \begin{subfigure}[t]{0.5\textwidth}
        \includegraphics[scale=0.28]{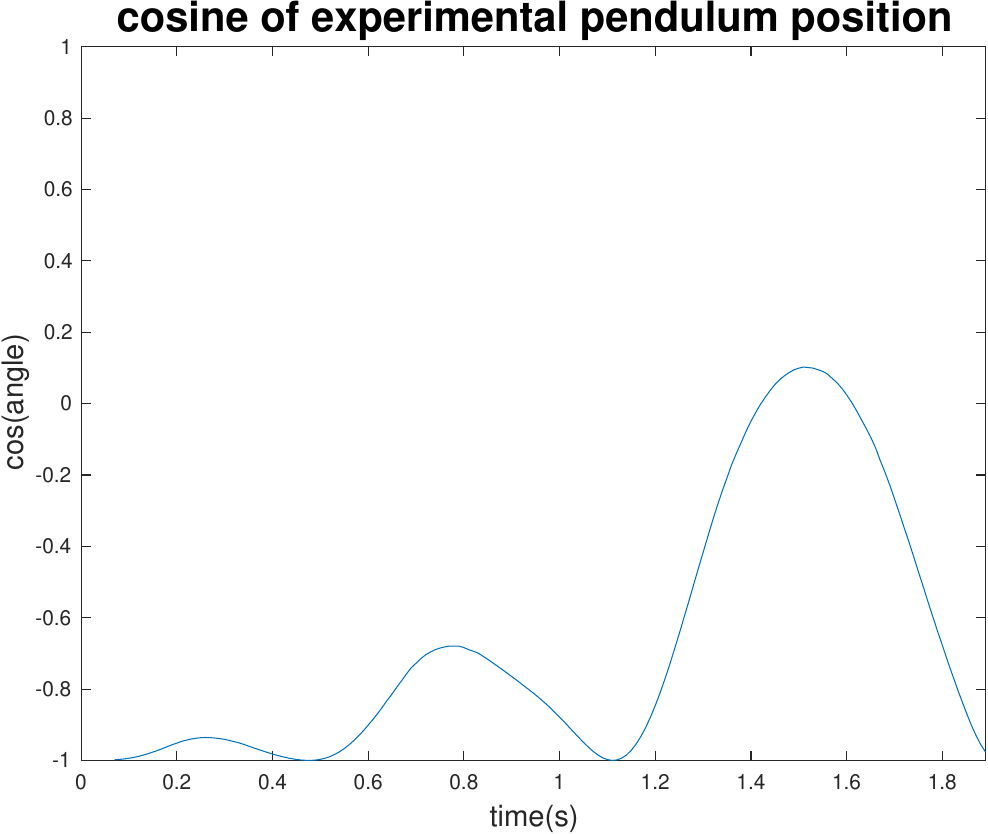}
        \label{fig:swing_old_cos}    
    \end{subfigure}%
    \hfill
    \begin{subfigure}[t]{0.5\textwidth}
        \includegraphics[scale=0.28]{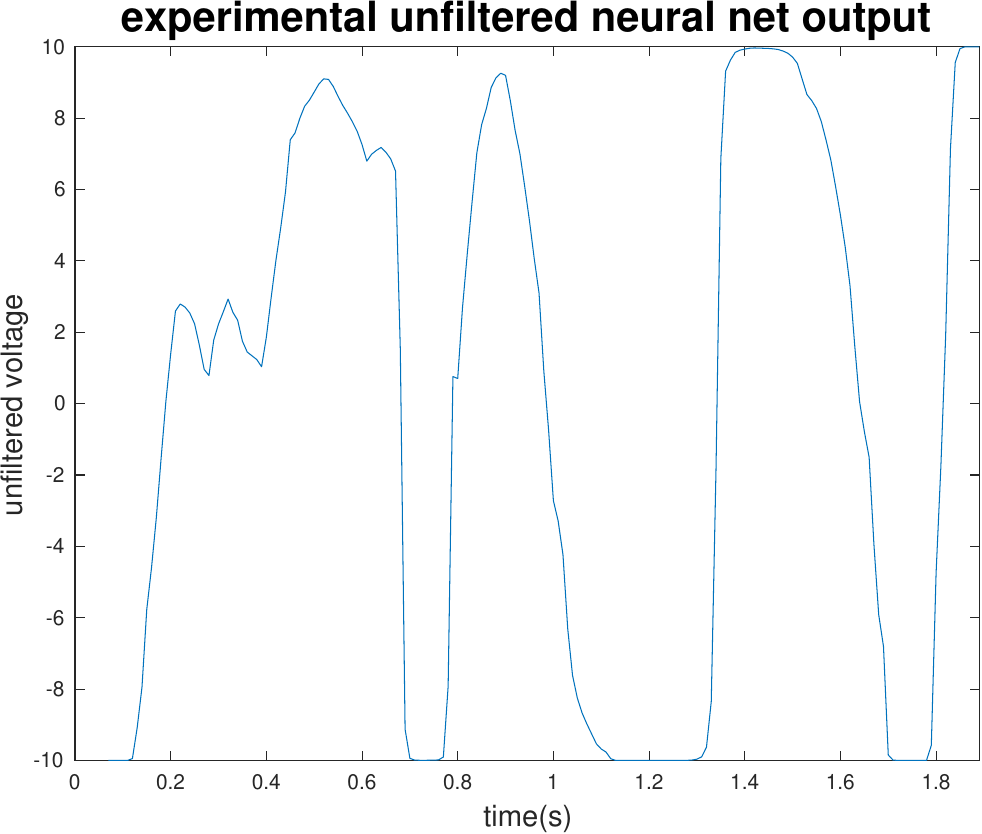}
        \label{fig:swing_old_vm}    
    \end{subfigure}
    \caption{First controller damaged the system in less than 2 seconds. In bottom right plot, we can see the control inputs nearly switched between saturated -10 and 10 volts in less than 0.05 second both around $t=0.7s$ and then again around $t=1.3s$}
    \label{fig:swing_old_damage_lab}
\end{figure}

An initial training following a similar approach as described in the previous section yielded a controller that was shown to be successful in the swing-up task in a deterministic simulation environment, but fails catastrophically in the lab. In particular, the control signal displayed bang-bang behavior, causing the cart to slip when the its direction was being changed, and subsequently broke the plastic wheels. Figure~\ref{fig:swing_old_damage_lab} shows the recorded trajectory of such an instance, where we had to stop the experiment after about 2 seconds due to the damaged cart wheel. 

\begin{figure}
    \begin{subfigure}[t]{0.49\textwidth}
    \includegraphics[scale=0.30]{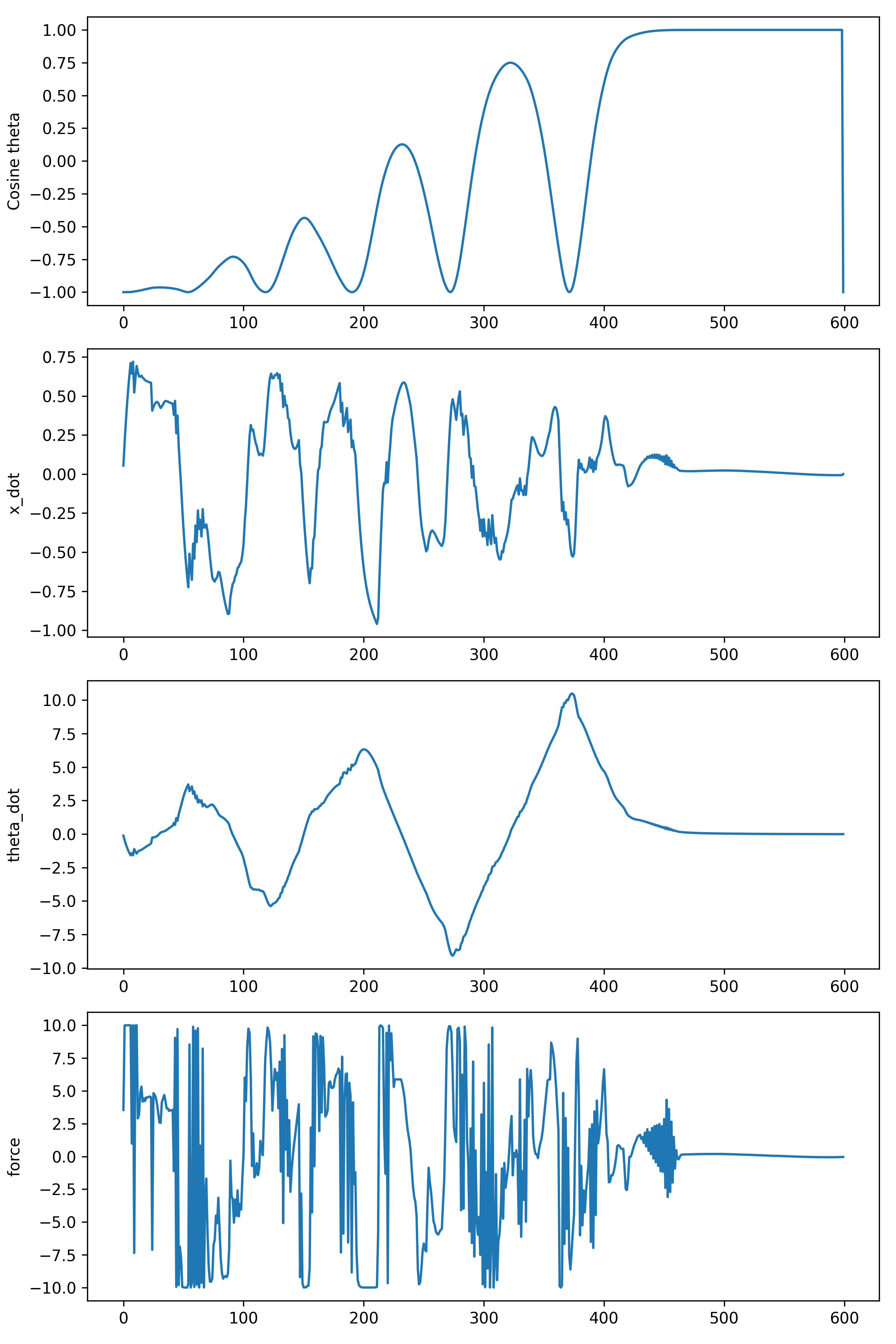}
        \caption{Controller trained with no uncertainty in model parameters, simulated with no noise}
        \label{fig:swing_old_no_noise}    
    \end{subfigure}%
    \hfill
    \begin{subfigure}[t]{0.49\textwidth}
    \includegraphics[scale=0.30]{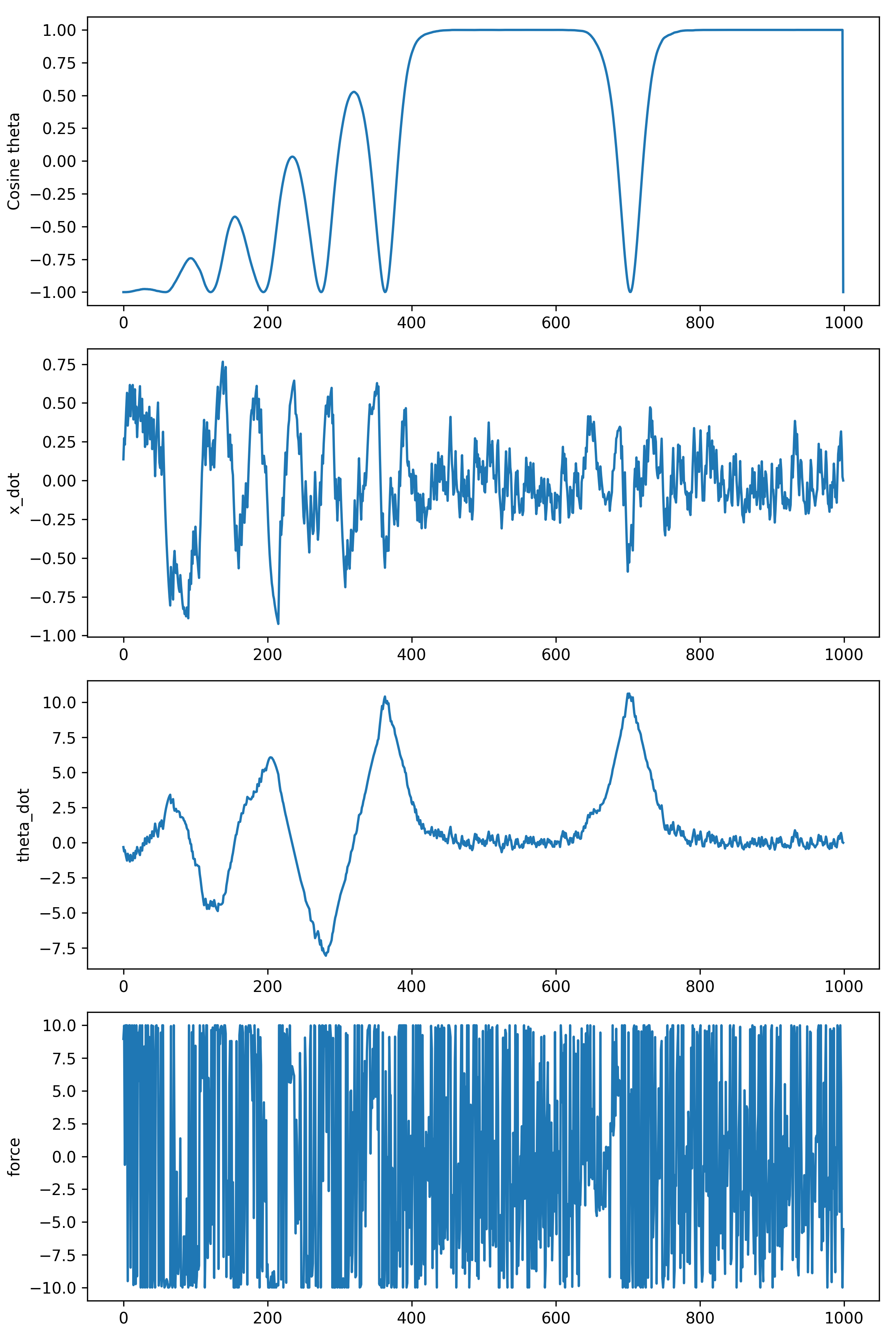}
        \caption{Controller trained with no uncertainty in model parameters, simulated with $\mathcal{N}(0,0.05)$ state and action noise} 
        \label{fig:swing_old_with_noise}    
    \end{subfigure}
    \caption{First controller trained with no uncertainty tends to saturate control input and switches violently. The first row of each plot is the cosine of the pendulum angle; the second row is the cart velocity; the third row is the pendulum's angular velocity; the fourth row is the control input voltage to the system.}
\end{figure}

The recorded voltage supplies were oscillating drastically in two particular instances, as summarized in Table~\ref{tab:broken_wheels}. 
The average rates of change in those instances both exceeded 750 V/s and created too much jerky movements that caused the wheel to slip and break. However, the swing-up motion appeared to work as intended before the wheel broke, which shows promise. We postulate that signal noise and model mismatch caused the controller trained in a noiseless environment to fail at such a complex task where the model parameters may depend on time or states. We then simulated the controlled system by adding moderate noise, $\mathcal{N}(0,0.05)$, to the states and control, the result is shown in Figure~\ref{fig:swing_old_with_noise} which corroborates with our lab observation of violent control input switching. 

\begin{table}[htbp]
    \centering
    \begin{subtable}{0.45\textwidth}
        \centering
        \begin{tabular}{|c|c|}
            \hline
            Time (s) & $V_m$ (V) \\
            \hline
            0.67 & -6.5192 \\
            \hline
            0.68 & 1.5239 \\
            \hline
            0.69 & 9.1256 \\
            \hline
        \end{tabular}
        \caption{Jerky cart movement \#1}
        \label{tab:wheelie1}
    \end{subtable}%
    \begin{subtable}{0.45\textwidth}
        \centering
        \begin{tabular}{|c|c|}
            \hline
            Time (s) & $V_m$ (V) \\
            \hline
            1.33 & 8.3478 \\
            \hline
            1.34 & 1.7838 \\
            \hline
            1.35 & -6.8664 \\
            \hline
        \end{tabular}
        \caption{Jerky cart movement \#2}
        \label{tab:wheelie2}
    \end{subtable}
    \caption{Rapid switching of control inputs without low-pass filter.}
    \label{tab:broken_wheels}
\end{table}

Given these initial observations, we believe that the RL algorithms being agnostic to the motor's physical limitations as well as the minor model mismatches during the swing-up trajectory may be the main causes for the zero-shot policy transfer to fail. Furthermore, even though the rapid oscillations of the controller actions are in some sense closer to being mathematically optimal, in reality they also tend to cause more wear and tear to the lab equipment. This motivates us to implement a discrete first-order low-pass filter on the action during training to filter out the high-frequency control signals and promote smooth actions. In addition, we also introduce randomization to the model parameters to avoid overfitting to the nominal environment in case unmodeled dynamics or other noise and disturbances occur. To identify the set of parameters that are suitable for domain randomization, we first perform a sensitivity analysis. 

\section{Sensitivity Analysis}
\label{sec:sensitivity}
We analyze the sensitivity of model parameters of the uncontrolled system in Equations~\ref{eq:lab_dynamics}. 
Given a dynamical system:
\begin{equation}
\label{eq:general_IVP}
    \dv{\vec{z}}{t} = \vec{f}(t, \vec{z}(t), \vec{p}), \quad \vec{z}(t_0) = \vec{z}_0, 
\end{equation}
define a (normalized) sensitivity matrix $\vec{s}_{jk}$ for the $j$-th state variable $\vec{z}_j$ with respect to the $k$-th parameter $\vec{p}_k$ near the nominal parameter values $\vec{p}^*$ as
\begin{equation}
\label{eq:sensitivity_matrix}
    \vec{s}_{jk}= \parenthesis{\pdv{\vec{z}_j}{\vec{p}_k} } \cdot (\vec{p}^*) . 
\end{equation}
To approximate the partial derivatives in (\ref{eq:sensitivity_matrix}), we use a complex-step method \cite{lyness_numerical_1967, squire_using_1998, martins_automated_2000, banks_complex-step_2015} as shown in Equation~\ref{eq:complex_diff}:
\begin{equation}
    \label{eq:complex_diff}
    \pdv{\vec{z}_j}{\vec{p}_k} \Bigg|_{\vec{p}^*} \approx 
    \frac{\textbf{Im}(\vec{z}_j(\vec{p}^* + ih \vec{e}_k ))}{h}.
\end{equation}
Here, $h>0$ is the step size; $\vec{e}_k$ is the standard $k$th unit vector in Euclidean space with a 1 at the $k$th coordinate and 0 elsewhere; and $i = \sqrt{-1}$; and $\textbf{Im}(\cdot)$ denotes the imaginary part of the complex-valued function. The complex-step method enjoys superior numerical stability for analytic functions, i.e. functions that are locally representable by a convergent power series. Note that by the analytic assumption, the step size can be reduced to machine precision and usually has negligible effect on the accuracy of the approximation \cite{banks_complex-step_2015}. 

For uncontrolled systems, we first compare our complex-step approximation of each sensitivity matrix with the Jacobians computed using reverse-mode auto-differentiation in JAX \cite{bradbury_jax_2018}. For three fixed initial conditions, the RMSE of each sensitivity matrix trajectories are less than $10^{-7}$. 
We also noticed a significant improvement in computational speed using our complex-step method in all the simulation scenarios that we experimented with, reducing computation time by as much as 5 times compared to JAX. 

Under some boundedness assumptions, it has been shown that gradient-based sensitivity measures can also be used to bound global variance-based sensitivity indices \cite{sobol_derivative_2009,kucherenko_derivative-based_2016,alexanderian_variance-based_2020}.
For a scalar function $f$ whose output depends on some uncertain parameters, $\vec{p}$, as well as on time, $t$, i.e.,
\[
Y = f(t, \vec{p}), \quad t \in [0,T], 
\]
where $\vec{p} \in \R^m$ is a vector of uncertain model parameters, quantifying the impact of $\vec{p}$ on $Y$ is highly dependent on efficient sampling and often incurs prohibitively high computational cost. The total effect index, denoted by $S_k^{tot}$, gives the total variance in $Y$ caused by the $k$-th parameter, ${p}_k$, as well as its interactions with any of the other parameters. This index is computed as 
\[
S_k^{tot} = 1 - \frac{\mathbb{V}[\E[Y|\vec{p}_{\sim k}]]}{\mathbb{V}[Y]}, 
\]
where $\vec{p}_{\sim k}=(p_1, \dots, p_{k-1}, p_{k+1}, \dots, p_m)$ denotes all the parameters except $p_k$; $\mathbb{V}[\cdot]$ and $\E[\cdot]$ denote the variance and expectation operators, respectively. 

\begin{figure}
    \centering
    \includegraphics[width=0.9\linewidth]{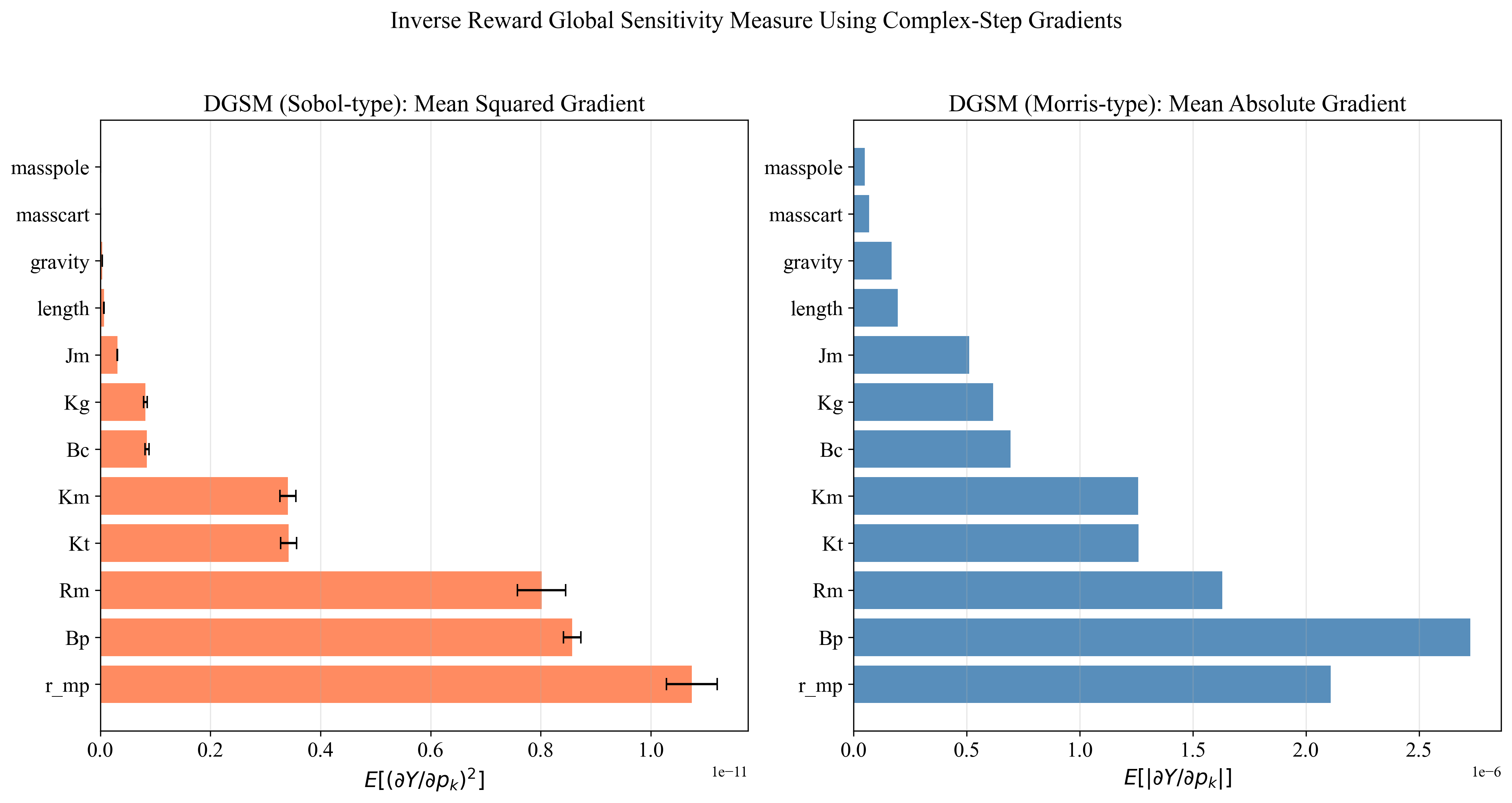}
    \caption{Derivative-based Global Sensitivity Measures (DGSM)}
    \label{fig:sensitivity-grad}
\end{figure}

More specifically, suppose that each model parameter $\vec{p}_k$ is uniformly distributed on the unit hypercube, and suppose that $f$ is a scalar function of a state variable, $\vec{z}_j$, of our dynamical system for a fixed initial condition, and further assume that $\pdv{\vec{z}_j}{\vec{p}_k}$ is square-integrable for each $j,k$, then we can obtain a practical estimation of the values of the total Sobol index of the parameter $k$ for variable $j$, using the lower and upper bounds based on the gradient method \cite{kucherenko_derivative-based_2016}. Therefore, we apply the complex-step method to approximate the partial derivatives and simulate the sensitivity of each model parameters in an uncontrolled free-fall simulation of the cart-pole system. We used the inverse of the cumulative reward of 20-second episodes with the pendulum starting near the upright position as the objective function, $Y$, for global sensitivity analysis. The parameters of cart mass, pole mass and length are assumed to have a 1\% uncertainty around their nominal values, gravity is assumed to vary between 9.75 and 9.9 $m/s^2$, while all other parameters are assumed to have 50\% uncertainty. The parameters are rescaled so that the uncertainty domain is a unit hypercube. 10,000 points were sampled on the hypercube to estimate the mean and variance. The results of the gradient-based sensitivity indices are shown in Figure~\ref{fig:sensitivity-grad}. 

Global sensitivity analysis (GSA) is a powerful tool for uncertainty quantification, but it often comes with a substantial computational overhead, especially for complex systems. 
In some recent work, GSA has been integrated into model-based RL with great promise \cite{polydoros_survey_2017,ramos_bayessim_2019,chatzilygeroudis_survey_2020}. 
In our project, the main objective is to identity a subset of model parameters for domain randomization (DR), while essentially preserving the dynamics of the nominal system. 
The rationale is that, since our stabilization controller has been trained in an uncertainty-free simulation environment and then successfully transferred to the physical system in a zero-shot manner, we are confident that our model in Equation~\ref{eq:lab_dynamics} is close enough to reality that it requires little modification. 
Unlike many DR applications where the environments need to be randomized to adapt to a wide range of operating conditions for the robots, our lab provides a stable environment so that our cart-pole system consistently operates close to the nominal model. 
We observe with some case studies in the next section that one needs to pay close attention to which parameters to randomize and by how much. 
In one case, we found that while randomizing three parameters resulted in a satisfactory policy, adding one more parameter to randomize negatively affected the training result. 

\section{Case Study}
\label{sec:case_study}

In this section, we compare the control performance of a few cases of using various combinations of the following training techniques: low-pass filter, domain randomization, and curriculum learning. 

In light of the high-frequency control feedback when the states are influenced under minor noise, we first apply a low-pass filter to the control signal that acts as a smoothing agent. 
Recall the following definition of low-pass filter in classical signal processing:
    Let the input signal be $x \in \R$, let the filtered output signal be $y$. A low-pass filter can be modeled by 
    \begin{equation}
        \tau \dot{y} + y = x, 
    \end{equation}
    where $\tau>0$ is the filter time constant. Its transfer function is
    \begin{equation}
        H(s) = \frac{1}{\tau s + 1},
    \end{equation}
    and the cut-off frequency is $1/\tau$. In the discrete setting with fixed step size $h$, the filtered control signal at time $t$ can be computed as
    \begin{equation}
        y_{t+1} = y_t + \frac{h}{\tau}(x_t - y_t), 
    \end{equation}
    where $y_t$ and $y_{t+1}$ are the filtered signals at time $t$ and $t+1$, respectively, and $x_t$ is the input of the raw signal at time $t$. 

During RL training, this filter is implemented as the discrete low-pass filter with the integration step size being the same as the environment. In the lab, this filter is applied to the output signal of the swing up policy. In both cases, we chose the filter constant $\tau=0.3$. %
This value was chosen relatively high as an attempt to minimize the jerky movements and to protect our lab equipment, but this value can be tuned and improved for better performance. 

In addition to action smoothing, we also apply domain randomization to diversify the training environment to further improve the sim-to-real transfer performance. Specifically, upon each episode reset, some model parameters are randomized within some predetermined intervals instead of being fixed across all simulation episodes. 
Introducing domain randomization comes with higher computational cost and slower convergence. To address this problem, we also adopt a curriculum learning schedule that slowly increases the level of model uncertainty during training. 
In our project, we adopt an episodic curriculum learning strategy with a linear schedule. That is, for a fixed number of initial episodes, the uncertainty margin for each randomized parameter increases linearly from zero to a specified maximum value. 

During training, each randomized parameter is uniformly sampled at the beginning of an episode from an interval around its nominal value, determined by the current curriculum margin. The value of the sampled parameter remains fixed for the duration of the episode. After reaching the maximum number of curriculum episodes, the uncertainty margin remains at its maximum, and all subsequent training episodes use parameters randomized at this maximum margin. This approach enjoys the benefit of stabilizing initial training due to reduced initial stochasticity, as well as potentially leading to better final performance. 

We study the effect of domain randomization and curriculum learning by comparing three training cases. For each case, training is terminated after 3 million simulation steps, and the neural network policies are recorded at every one-million step mark, as well as whenever the ``best'' episodic rewards are achieved during training. 
To evaluate the performance of the controllers without risking any damage to the lab equipment, we simulate each controlled system under various levels of signal noise and model uncertainty, and record how often and how quickly it swings the pendulum near the upright position. We evaluate each trained controller over 100 simulation episodes, each with 700 steps, under increasing levels of disturbance. The disturbance levels are summarized below in Table~\ref{tab:testing_combos}.

\begin{table}[htbp]
    \centering
        \begin{tabular}{|c|c|c|c|}
            \hline
            Disturbance Level & state noise & action noise & parameter uncertainty margin \\
            \hline
            clean & 0 & 0 & 0.01\% \\
            \hline
            moderate & $\mathcal{N}(0,0.02)$ & $\mathcal{N}(0,0.01)$ & 0.05\% \\
            \hline
            severe & $\mathcal{N}(0,0.05)$ & $\mathcal{N}(0,0.02)$ & 0.10\% \\
            \hline
        \end{tabular}
        \caption{Testing trained scenarios with various levels of disturbances}
        \label{tab:testing_combos}
\end{table}

The evaluation statistics of the simulated trajectories are compiled into tables and each table consolidates the following information.
\begin{enumerate}
    \item The first column of each table specifies a level of disturbance (rows in Table~\ref{tab:testing_combos}). 
    \item The second column documents the average episodic reward for 100 simulated trajectories in the corresponding noise level. 
    \item The third column documents the standard deviation of the episodic rewards; the fourth column documents the average time the pendulum is near the upright position, constrained by the following:
    \begin{itemize}
        \item the cart position is within 0.4 meter from the center, which is dangerously close to the boundary of the track, but it is less important than the pendulum angle.
        \item the magnitude of the cart velocity is less than 3 $m/s$. 
        \item the pendulum angle is within 15 degrees (or $\pi / 12$ radian) from vertical, and 
        \item the magnitude of angular acceleration is less than $5\pi/6~ rad/s$.  %
    \end{itemize}
    We take the first instance where the system enters this region as the ``reach time''. If this entry is empty, it means that the system did not enter the near upright region within the total simulation time of 7 seconds (700 steps). 
    \item The last column documents how many of the simulated trajectories reached the near upright region.
\end{enumerate}

\subsection{Case 0: Randomize all physical parameters with 10\% margin over 100 episodes}

A naive approach is to randomize all physical parameters without additional consideration. In this case, whenever domain randomization is enabled, we randomly randomized all physical parameters and each parameter is sampled uniformly within an uncertainty margin of up to 10\% near their nominal values; whenever curriculum learning is enabled, the randomization was structured by a linear curriculum, which schedules the uncertainty margins of each parameter to increase from 0 to 10\% over the course of 100 episodes. Under these specifications, we trained 3 controllers using 3 training configurations and tested their control performances according to the disturbance levels listed in Table~\ref{tab:testing_combos}. The three training configurations are: (i) training with both domain randomization and curriculum learning, (ii) training with domain randomization but without curriculum learning, and (iii) no domain randomization. We use the control policy recorded with the highest episodic reward for this evaluation. The evaluation results are summarized in Table~\ref{tab:test_case0}. 

\begin{sidewaystable}
\caption{Result of controllers obtained from case 0 of randomizing all model parameters, trained with or without curriculum learning, simulated under increasing level of disturbance}
\label{tab:test_case0}
    \centering
        \begin{tabular}{|c||l||c|c|c|c|}
            \hline
            \multicolumn{2}{|c|}{} & \multicolumn{4}{c|}{Data columns} \\
            \hline
            \multicolumn{1}{|c|}{} 
            & Disturbance Level & reward mean & reward std & avg reach time (s) & reach rate \\
            \hline\hline
            \multirow{3}{*}{Enable DR, Enable CL} 
            & clean & -1330.94 & 7.50 & 2.06 & 100/100 \\
            \cline{2-6}
            & moderate & -1475.06 & 313.05 & 2.16 & 100/100 \\
            \cline{2-6}
            & severe & -2352.85 & 306.28 & 2.58 & 88/100 \\
            \hline
            \hline
            \multirow{3}{*}{Enable DR, Disable CL}
            & clean & -1454.23 & 102.77 & 2.46 & 100/100 \\
            \cline{2-6}
            & moderate & -1591.69 & 211.88 & 2.63 & 100/100 \\
            \cline{2-6}
            & severe & -2154.86 & 260.92 & 2.88 & 0/100 \\
            \hline
            \hline
            \multirow{3}{*}{Disable DR, Disable CL}
            & clean & -1334.35 & 6.69 & 2.08 & 100/100 \\
            \cline{2-6}
            & moderate & -1576.33 & 249.51 & 2.25 & 100/100 \\
            \cline{2-6}
            & severe & -2322.22 & 308.20 & 2.42 & 83/100 \\
            \hline\hline
        \end{tabular}

\end{sidewaystable}

It is interesting that in such a highly stochastic training environment, the effect of domain randomization and curriculum are blurred. For example, while the average rewards are lower for small to moderate noise with curriculum enabled, this trend does not hold for the case of higher disturbance. Moreover, the difference between training configurations 1 and 3 appears to be insignificant if we compare their respective results in Table~\ref{tab:test_case0}. 
In theory, the performance of a controller trained on an environment with high variance should perform worse than the one learned with a high fidelity model. We believe that this warrants further investigation. 

\subsection{Case 1: Randomize $J_m$, $K_t$ and $K_m$ with 5\% margin over 100 episodes}

In Case 1, domain randomization is always enabled, and in particular, we randomly select three parameters based on the sensitivity analysis in Section~\ref{sec:sensitivity}, and the randomization is structured by a linear curriculum that increases each uncertainty margin from 0 to 5\% over the course of 100 episodes. During the first 100 training episodes, each parameter is sampled within the scheduled uncertainty margin at the beginning of the episode. After 100 episodes, each random parameter is sampled uniformly within $\pm5\%$ of its nominal value.
Based on our observation from the sensitivity analysis, the following three parameters have low to moderate effects on the systems trajectory, and were chosen to be randomized during training according to a linear curriculum:
\begin{itemize}
    \item rotor moment of inertia, $J_m$; 
    \item motor torque constant, $K_t$; and
    \item motor back-EMF constant, $K_m$.
\end{itemize}

Testing this controller at the same disturbance levels listed in Table~\ref{tab:testing_combos} yielded promising results, as summarized in Table~\ref{tab:test_case1_0TTT}. In particular, this controller achieved a 100\% success rate in the simulations of the severity level of the disturbance. Regrettably, because this training script had hard-coded configurations, we were unable to modify the training settings to test the other training settings, such as deactivating CL. As a result, this subsection reports only one training scenario, with both DR and CL enabled, while disabling either training techniques were not implemented. 

\begin{table}[htbp]
    \centering
    \caption{Test controller performance from case 1 trained with configuration 1 (both DR and CL are enabled), simulation with various levels of disturbances}
    \label{tab:test_case1_0TTT}
        \begin{tabular}{|c|c|c|c|c|}
            \hline
            Disturbance Level & reward mean & reward std & avg reach time (s) & reach rate \\
            \hline
            clean & -1371.05 & 5.30 & 2.06 & 100/100 \\
            \hline
            moderate & -1387.39 & 10.70 & 2.07 & 100/100 \\
            \hline
            severe & -1746.54 & 221.96 & 2.17 & 100/100 \\
            \hline
        \end{tabular}
\end{table}

\subsection{Case 2: Randomize $J_m$, $K_t$, $K_m$ and $B_p$ with 10\% margin with 500 episodes}
\label{subsec:case2}

Next, we would like to see what would happen if we increased the uncertainty levels around four parameters but also increased the length of the curriculum. The only new parameter that we introduced into the randomization in this case was the viscous damping coefficient on the pendulum hinge $B_p$. 
In this case, whenever domain randomization is enabled, we randomize four parameters; each parameter is sampled uniformly within $\pm10\%$ of their nominal values; whenever curriculum learning is enabled, the randomization was structured by a linear curriculum which schedules the uncertainty margins of each parameter to increase from 0 to 10\% over the course of 500 episodes. 

\begin{figure}[htbp]
    \begin{subfigure}{\linewidth}
        \centering
        \includegraphics[trim={0cm 1.5cm 0cm 0cm},clip,scale=0.3]{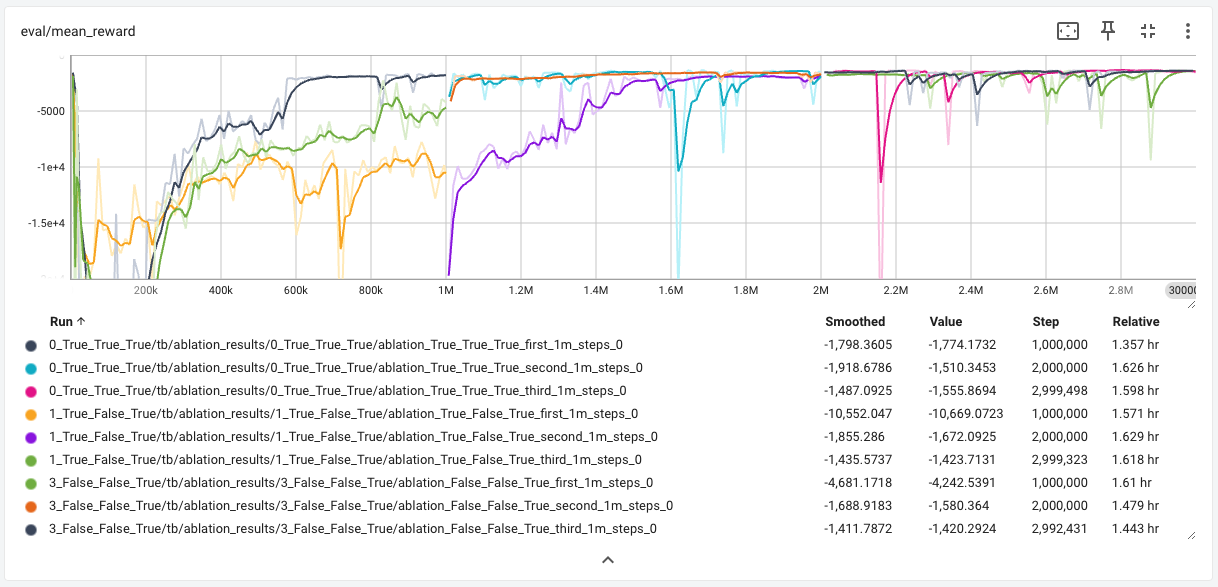}
        \caption{Training Reward History}
        \label{fig:training_curves_reward}
    \end{subfigure}
    \begin{subfigure}{\linewidth}
        \centering
        \includegraphics[trim={0cm 1.5cm 0cm 0cm},clip,scale=0.3]{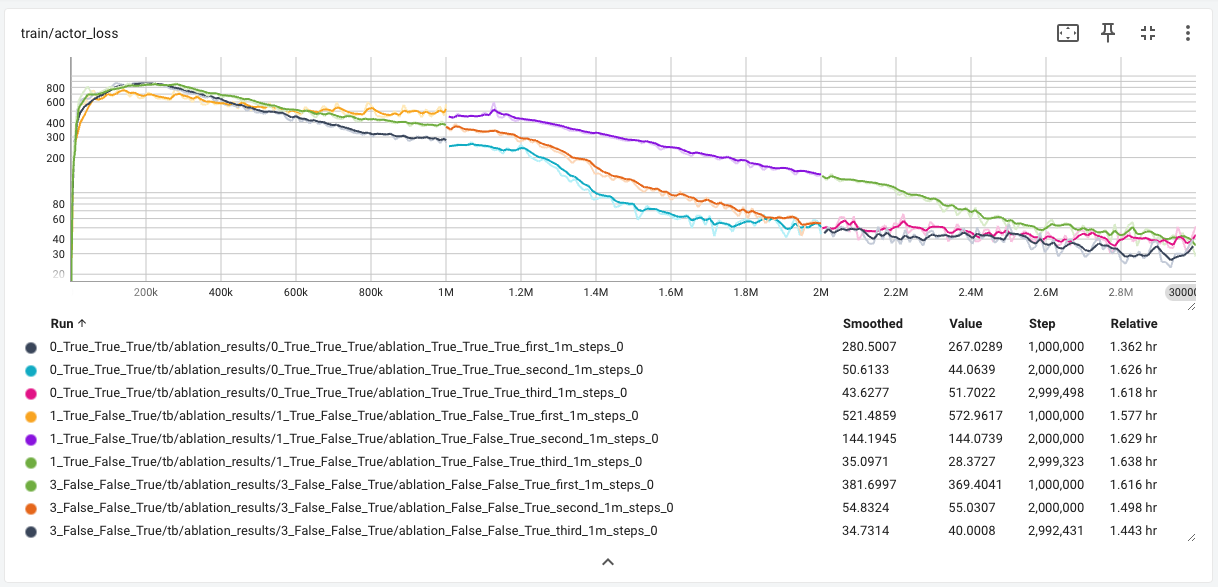}
        \caption{Training Actor Loss History}
        \label{fig:training_curves_actor}
    \end{subfigure}
    \begin{subfigure}{\linewidth}
        \centering
        \includegraphics[trim={0cm 1.5cm 0cm 0cm},clip,scale=0.30]{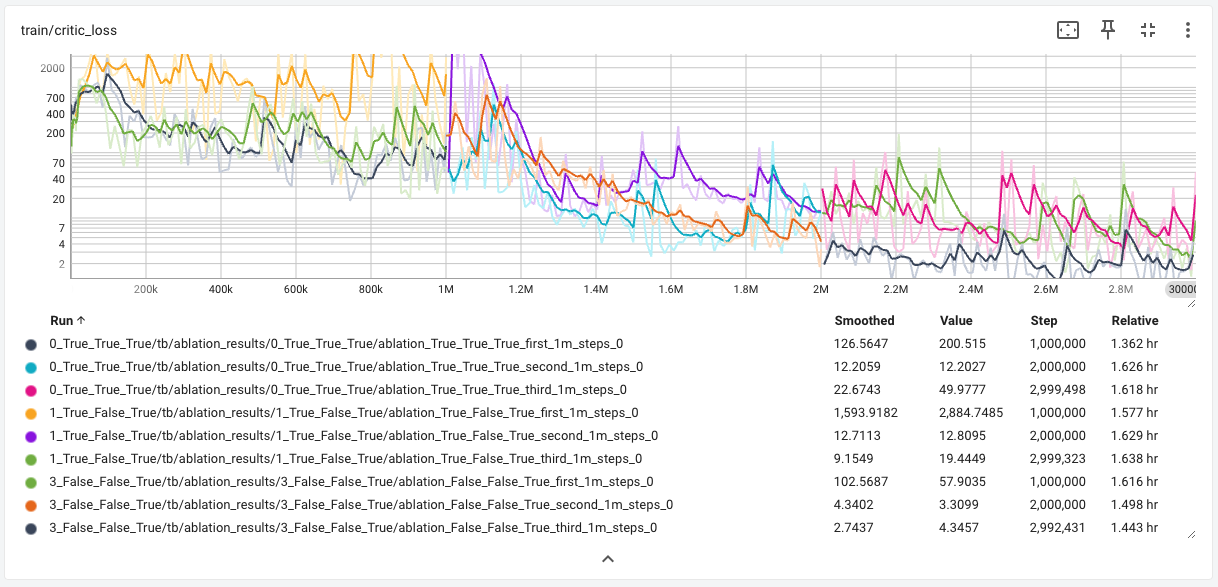}
        \caption{Training Critic Loss History}
        \label{fig:training_curves_critic}
    \end{subfigure}
    \caption{Screenshots of tensorboard training history with or without domain randomization and curriculum learning}
    \label{fig:dr+cl}
\end{figure}

Empirically, the combination of curriculum learning and domain randomization appears to be effective. As shown in Figures~\ref{fig:dr+cl}, both actor loss and critic loss decrease faster with curriculum learning enabled, while rewards increase faster with curriculum learning enabled. Similarly, without domain randomization, the actor loss is also lower, which is expected since there is less variance in the training system. However, we shall see later that when we apply the policies learned without domain randomization to the physical system, the controller fails to swing the pendulum anywhere near the upright position. Therefore, it is important to point out that training history alone may not be the correct metric to evaluate the effectiveness of a control policy, since any physical system will inevitably have noises and disturbances on top of modeling mismatches. 

The simulation results of Case 2 are summarized in Table~\ref{tab:test_case2}.
These results suggest that domain randomization and curriculum are beneficial both for stabilizing early training and for improving simulation performance. In particular, when CL was disabled while DR was enabled, the controller failed to complete the swing up task within 7 seconds. Meanwhile, with both CL and DR enabled, the average episode rewards were generally higher and the standard deviations were lower across the tested disturbance levels. Additionally, the average reaching times and reaching percentages were somewhat comparable, with a modest advantage for the controller trained with CL and DR at a higher level of disturbance. Note also that the curriculum schedule is longer in this case (500 episodes instead of 100); more study is still needed before drawing broader conclusions. 

\begin{sidewaystable}
\caption{Result of controllers obtained from case 2 of randomizing four model parameters, trained with or without curriculum learning, simulated under increasing level of disturbance}
\label{tab:test_case2}
    \centering
        \begin{tabular}{|c||l||c|c|c|c|}
            \hline
            \multicolumn{2}{|c|}{} & \multicolumn{4}{c|}{Data columns} \\
            \hline
            \multicolumn{1}{|c|}{} 
            & Disturbance Level & reward mean & reward std & avg reach time (s) & reach rate \\
            \hline\hline
            \multirow{3}{*}{Enable DR, Enable CL} 
            & clean & -1328.94 & 5.94 & 2.11 & 100/100 \\
            \cline{2-6}
            & moderate & -1517.15 & 319.83 & 2.16 & 100/100 \\
            \cline{2-6}
            & severe & -2326.58 & 261.05 & 2.50 & 88/100 \\
            \hline
            \hline
            \multirow{3}{*}{Enable DR, Disable CL} 
            & clean & -914.64 & 50.42 & - & 0/100 \\
            \cline{2-6}
            & moderate & -909.84 & 50.36 & - & 0/100 \\
            \cline{2-6}
            & severe & -911.55 & 52.99 & - & 0/100 \\
            \hline
            \hline
            \multirow{3}{*}{Disable DR, Disable CL} 
            & clean & -1341.43 & 8.13 & 2.09 & 100/100 \\
            \cline{2-6}
            & moderate & -1624.75 & 344.42 & 2.26 & 100/100 \\
            \cline{2-6}
            & severe & -2486.35 & 258.75 & 2.72 & 82/100 \\
            \hline
        \end{tabular}
\end{sidewaystable}

\section{Policy Switching}
\label{sec:swing_switch}

Since the upright equilibrium is unstable and using only the swing up controller for stabilization does not appear to be very robust, we decided to switch the control policies between the swing up and stabilization regimes in the lab. At first, our switching strategy was based on the previous region of attraction analysis for the robust stabilization controller; however, it turned out to be too restrictive, and there was still plenty of room on the track for the cart to maneuver. Secondly, the region of attraction was imposed with hard boundaries on the cart position and pendulum angle, but it was possible that those boundaries could be extended further and still retain stability. Additionally, improper switching can cause chattering or high frequency oscillation, which can cause actuator to deteriorate and destabilize the control system. Therefore, in practice, the switching strategy must be carefully designed to avoid these problems. 

\begin{figure}
    \centering
    \includegraphics[trim={0.1cm 1.5cm 4.5cm 2cm},clip,width=\linewidth]{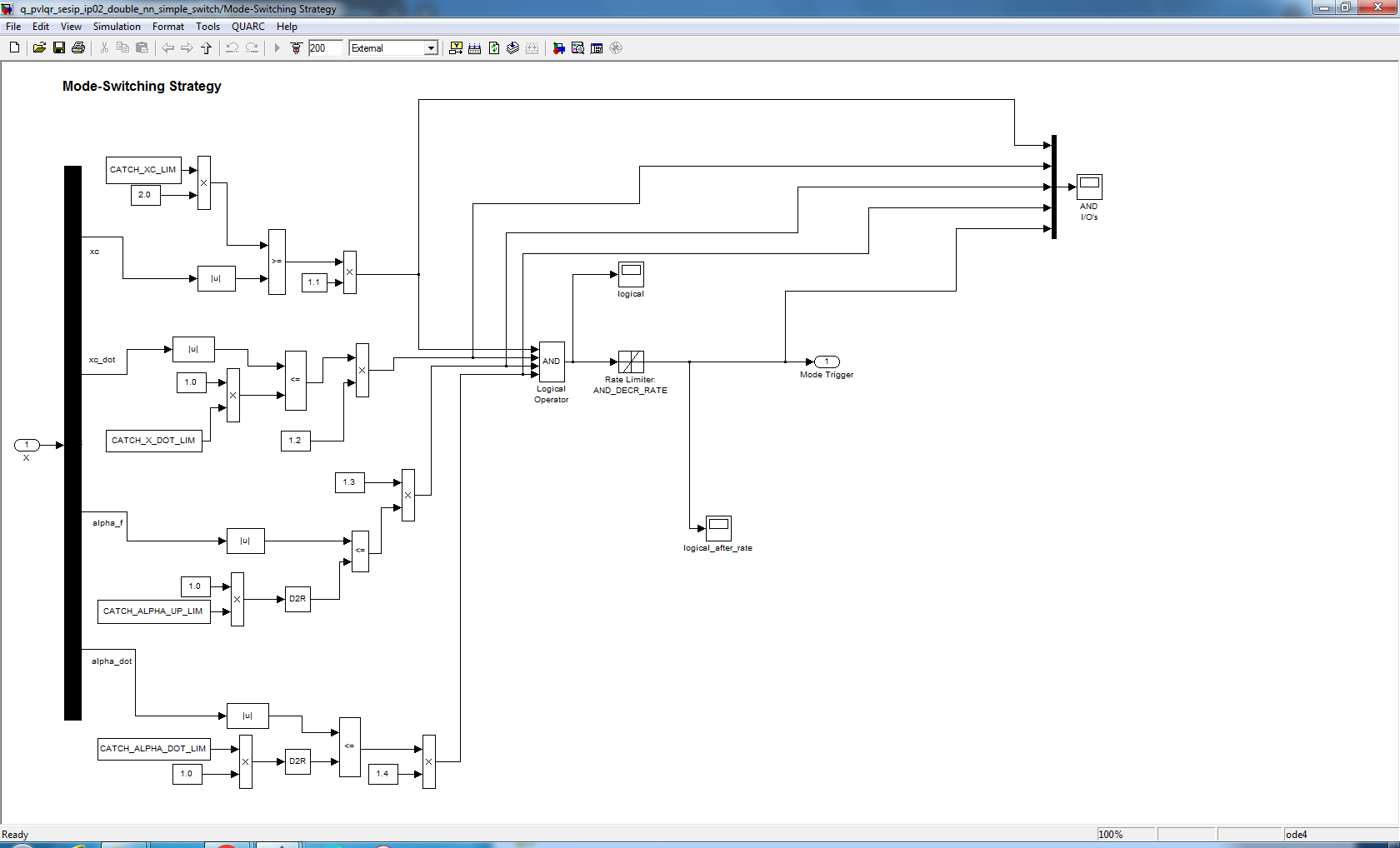}
    \caption{Simulink Switching Subroutine}
    \label{fig:simulink_switch}
\end{figure}

\begin{figure}
    \centering
    \includegraphics[trim={0.1cm 1.5cm 5.5cm 2cm},clip,width=\linewidth]{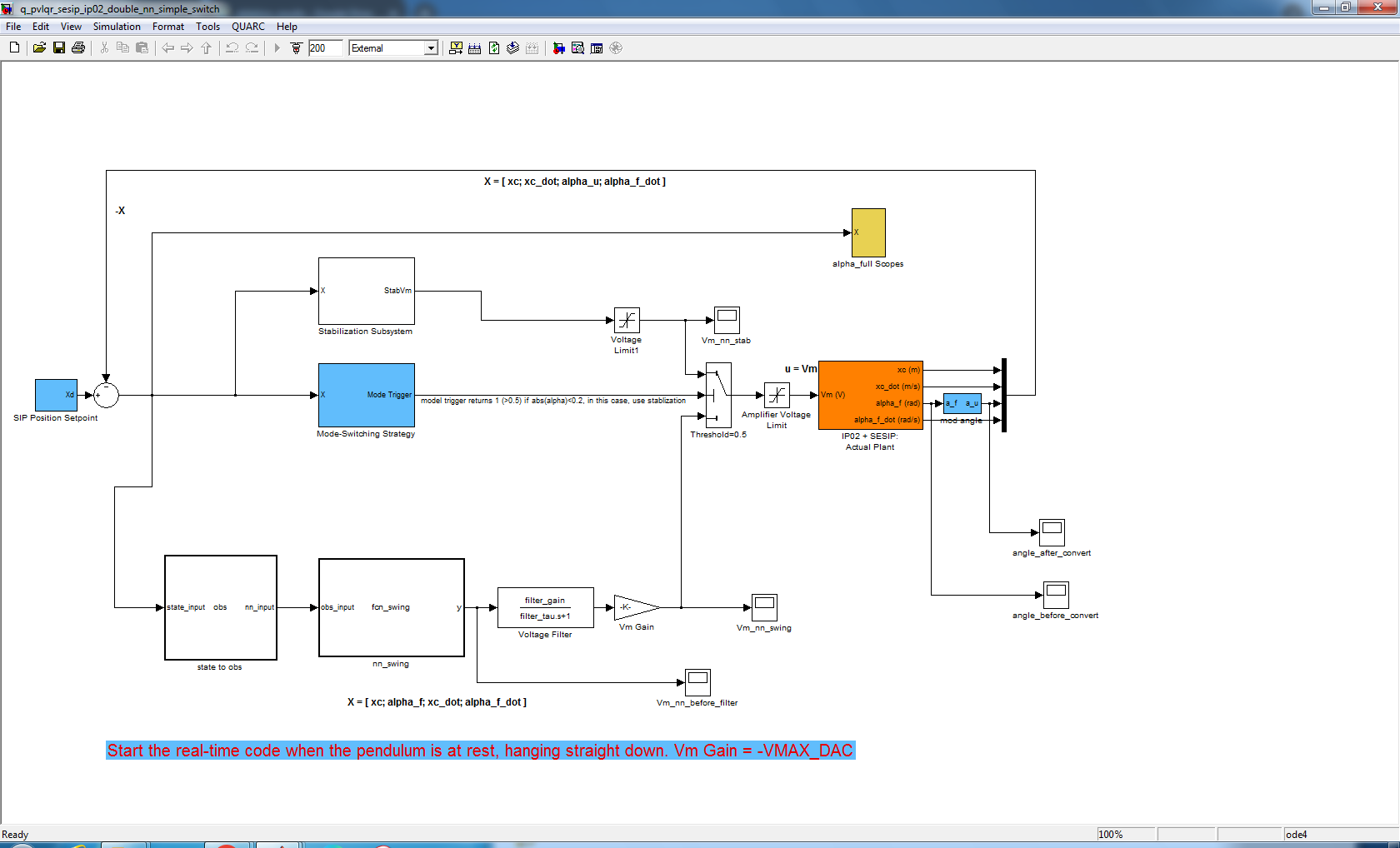}
    \caption{Overall Simulink Swing-up and Stabilization Implementation}
    \label{fig:simulink_final}
\end{figure}

After some tuning, we found that we could switch between the two regimes with satisfactory handoff behavior by enforcing a slightly slower drop rate than the entering rate and by using larger bounds on the state variables to increase the effective dwell time without imposing a hard minimum dwell time. The switched strategy implemented in Simulink is shown in Figure~\ref{fig:simulink_switch}. The general Simulink flow chart is shown in Figure~\ref{fig:simulink_final}. 

We tested all trained controllers in the lab, but the only successful controller was the one trained in Case 1, where randomization was applied only to the three low to moderately sensitive parameters. Increasing randomization to larger parameter sets leads to deteriorated performance in our current hardware setting. This may reflect a distribution shift caused by unrealistically broad uncertainty sets, but also indicates that the expected cumulative reward alone is not sufficient for controller selection. These observations are treated as empirical evidence from a limited campaign rather than as a definitive conclusion.

\begin{figure}[htbp]
    \begin{subfigure}{0.45\linewidth}
        \centering
        \includegraphics[scale=0.3]{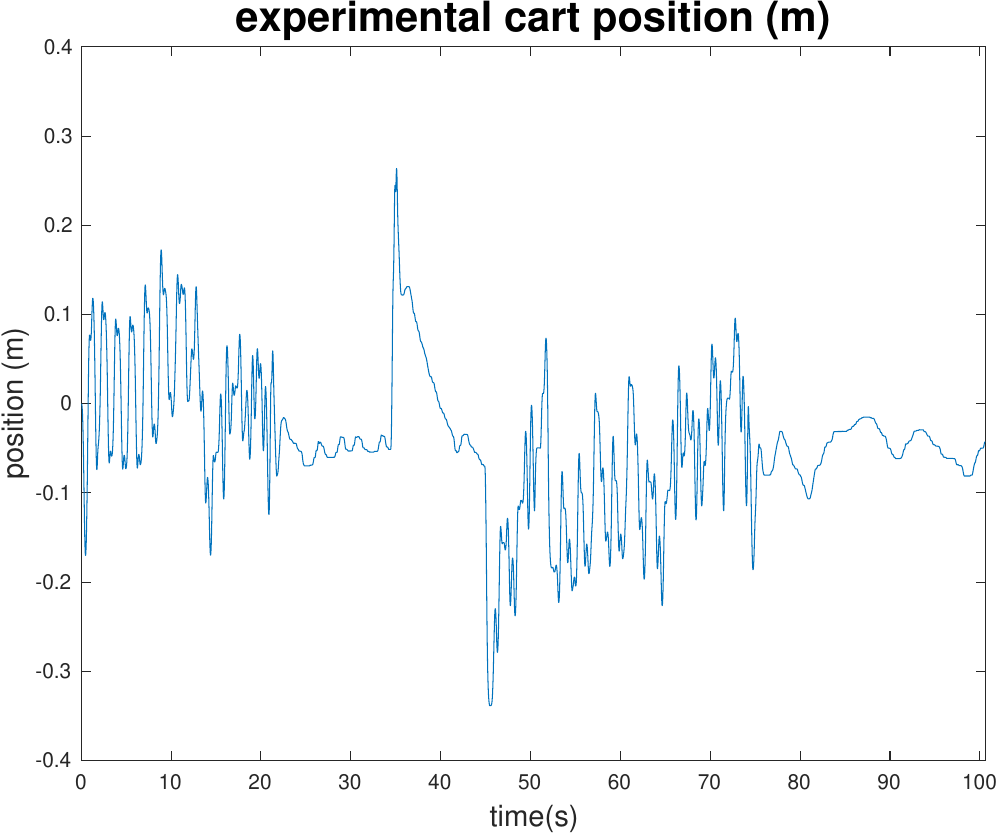}
        \label{fig:demo_x}
    \end{subfigure}%
    \begin{subfigure}{0.45\linewidth}
        \centering
        \includegraphics[scale=0.3]{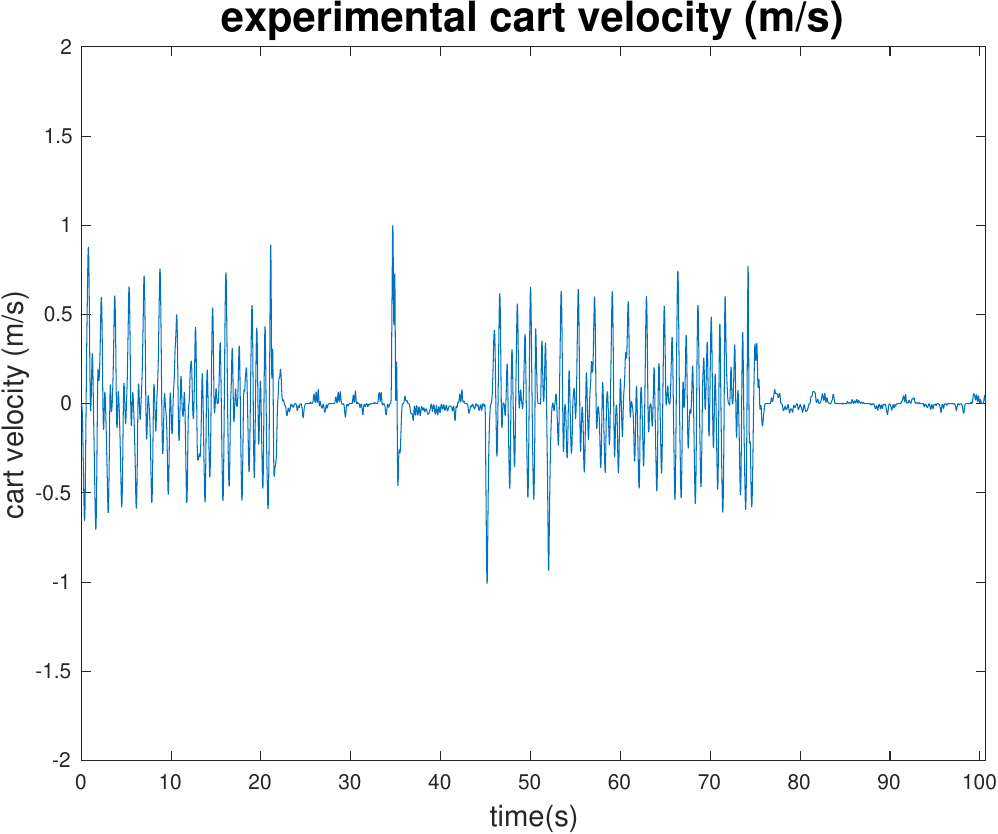}
        \label{fig:demo_xdot}
    \end{subfigure}
    \begin{subfigure}{0.45\linewidth}
        \centering
        \includegraphics[scale=0.3]{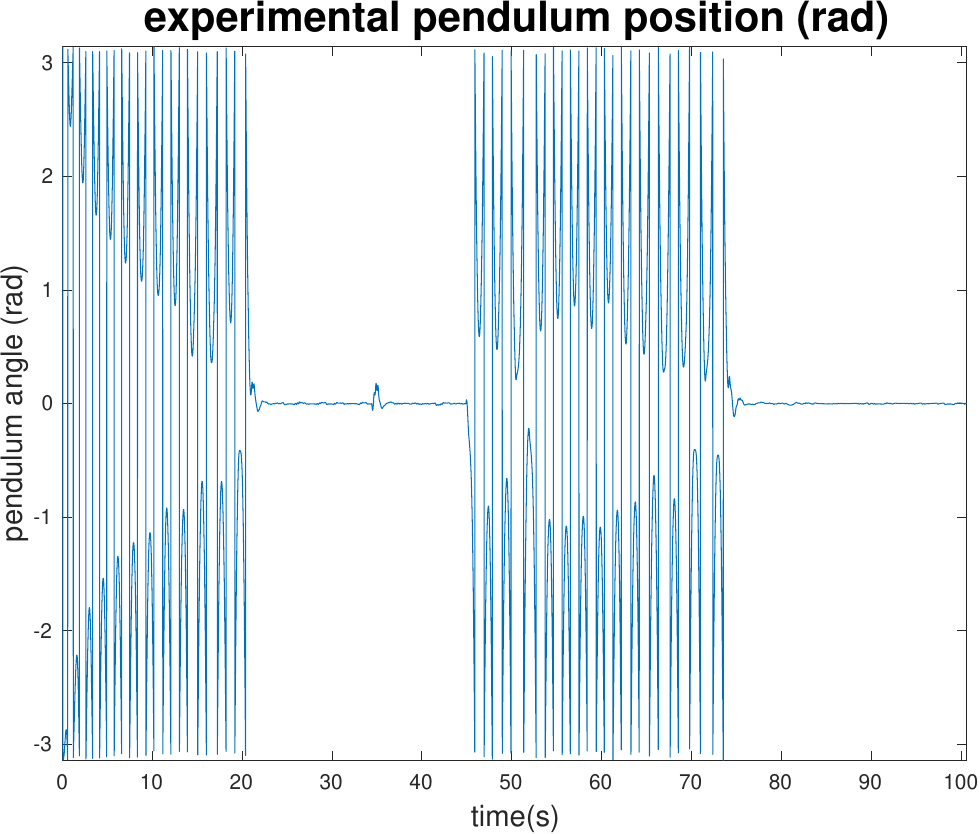}
        \label{fig:demo_alpha}
    \end{subfigure}%
    \begin{subfigure}{0.45\linewidth}
        \centering
        \includegraphics[scale=0.3]{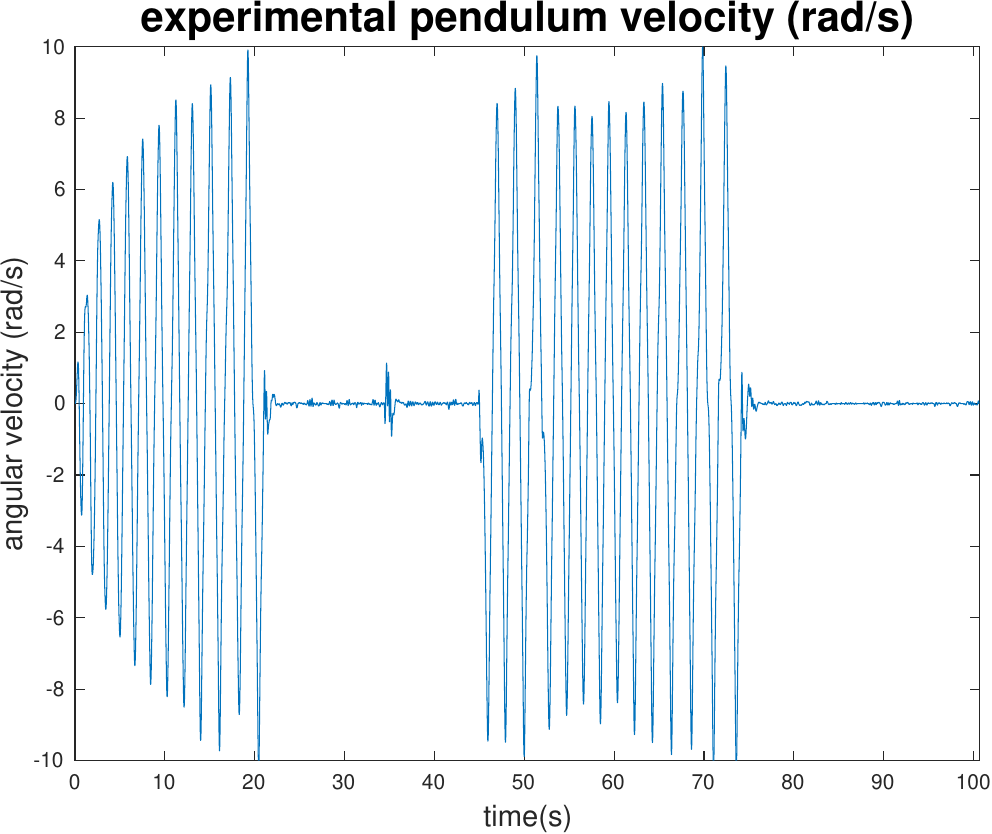}
        \label{fig:demo_alphadot}%
    \end{subfigure}
    \caption{Testing a Successful Control: a gentle tap around 35s showed the robustness of the stabilization control; another tap around 45s demonstrated recovery}
    \label{fig:demo}
\end{figure}

\begin{figure}[htbp]
    \begin{subfigure}{0.45\linewidth}
        \centering
        \includegraphics[scale=0.3]{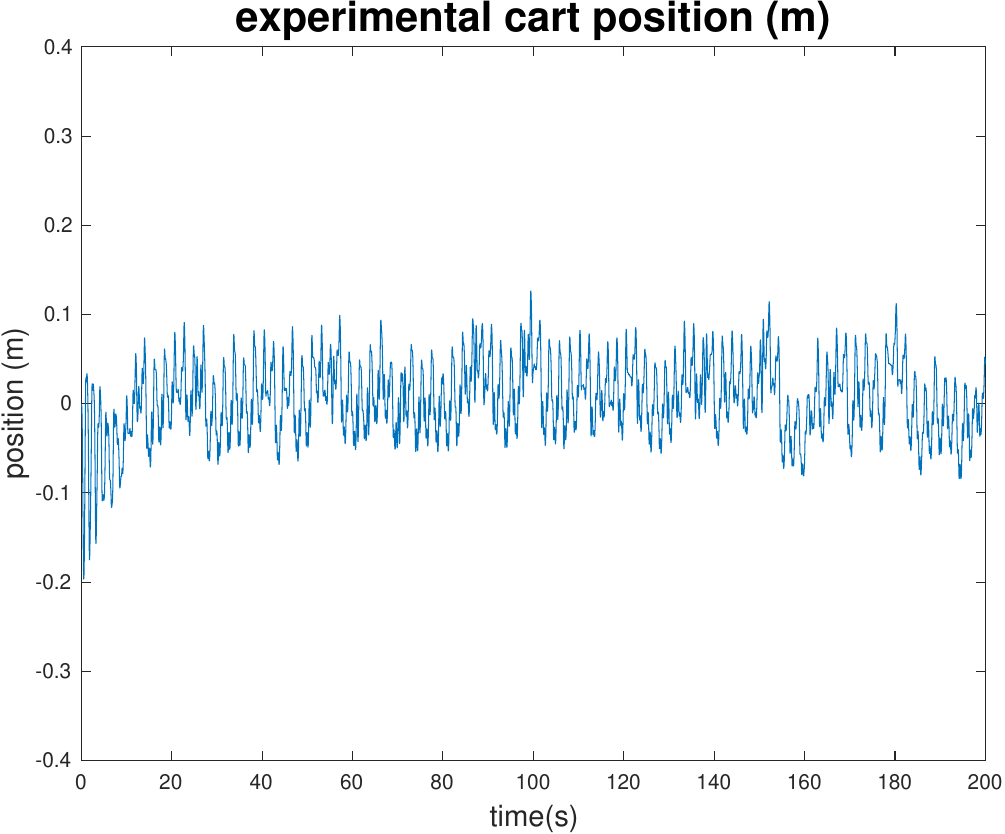}
        \label{fig:0TTT_x}
    \end{subfigure}%
    \begin{subfigure}{0.45\linewidth}
        \centering
        \includegraphics[scale=0.3]{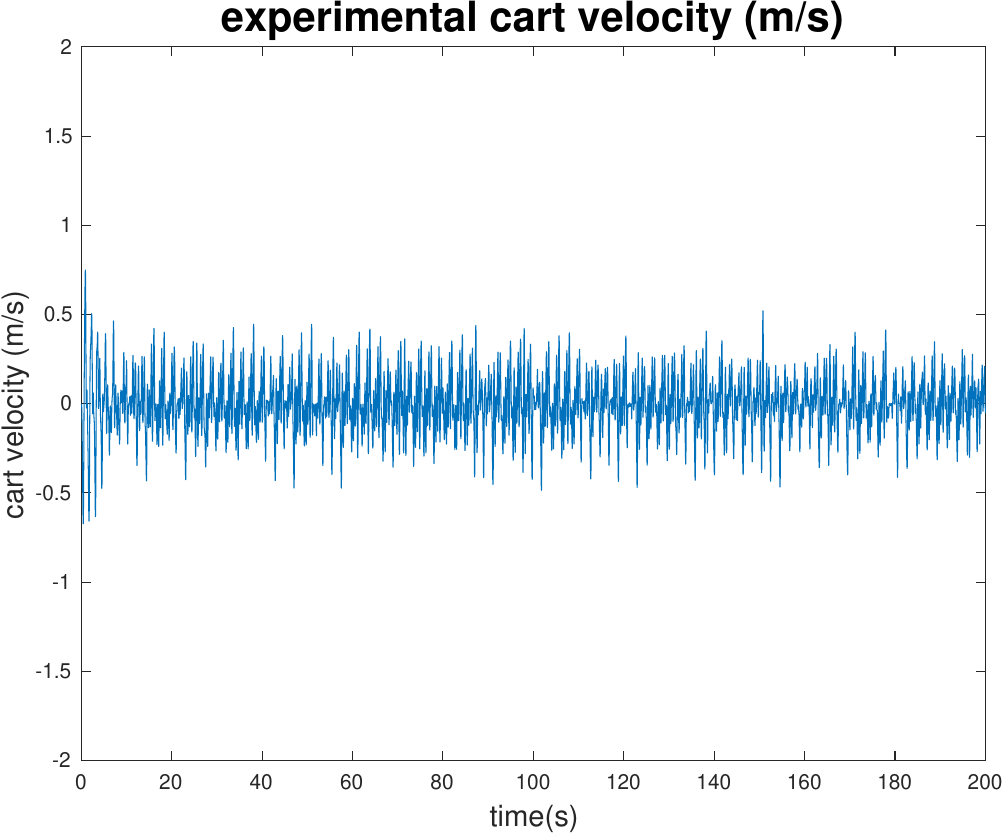}
        \label{fig:0TTT_xdot}
    \end{subfigure}
    \begin{subfigure}{0.45\linewidth}
        \centering
        \includegraphics[scale=0.3]{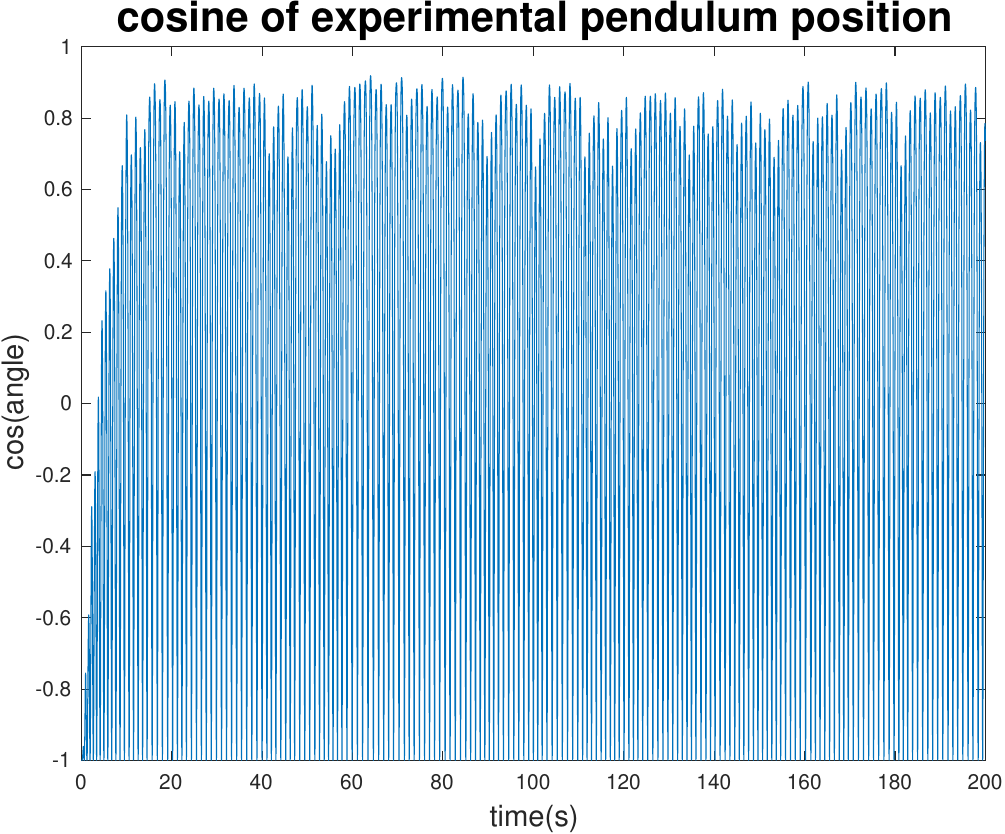}
        \label{fig:0TTT_alpha}
    \end{subfigure}%
    \begin{subfigure}{0.45\linewidth}
        \centering
        \includegraphics[scale=0.3]{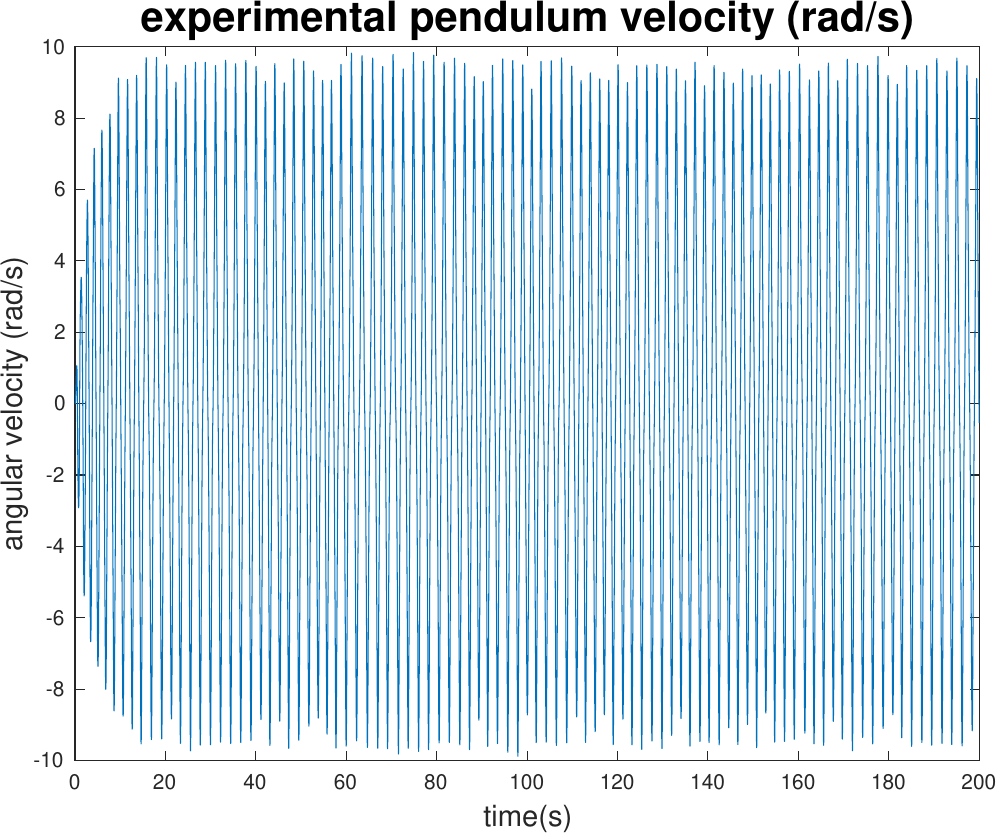}
        \label{fig:0TTT_alphadot}
    \end{subfigure}
    \caption{Testing a policy from Case 0 trained with domain randomization and curriculum learning: pendulum never reaches upright position}
    \label{fig:OTT_lab}
\end{figure}

In Figure~\ref{fig:demo}, we show a representative example of the switched switch from swing-up to stabilization control system. Although not very consistent, our control system typically takes about 15 to 40 seconds for the pendulum to go from downward to stable in the upright position. It can also recover itself if an external disturbance causes the pendulum to fall. However, other trained controllers that performed satisfactorily in the simulation failed to replicate that behavior in the lab. Figure~\ref{fig:OTT_lab} shows one such example, in which the pendulum remained swinging for 200 seconds without ever entering the switching region. 

\section{Discussion and Conclusion}

This paper presents a practical pathway towards obtaining a deployable RL-based control policy for the swing-up and stabilization of a cart-pole system. Under actuator limitations and noisy measurement signals, action filtering appears to be essential to protect hardware from high-frequency control inputs. Sensitivity-guided domain randomization provides a practical way to reduce overfitting to a nominal model, and incorporating curriculum learning helps stabilize early stages of training. We also provide an evaluation protocol to report the swing-up time, the success rate, in addition to the mean of episodic rewards, and the standard deviation. 

Our study empirically shows that training rewards alone may be a poor indicator of hardware performance; in most cases, controllers with higher simulated rewards did not transfer well to the lab. Our experiments also show that action filtering is a safety-critical component to our system, and the policies trained with the combination of targeted domain randomization and curriculum learning transfer are better than the policies trained with broader indiscriminate randomization. 
Moreover, switching logic is an equally important design consideration. For switched systems, hysteresis in the stabilization region and dwell time are important practical considerations \cite{liberzon_switching_2003}. In our lab implementation, we address these issues by relaxing the region for switching to cover the region of attraction of the stabilization controller, as well as adopting a slower deceleration rate for the logic signal upon state exit. 

Overall, our proposed hybrid RL-based control of the cart-pole system is as follows: action filtering, which respects the actuator bandwidth, is required to protect the hardware; domain randomization on a small set of model parameters guided by sensitivity analysis and a curriculum learning schedule can improve training performance; a switching logic that handles the handoff between two control policies should tolerate the hysteresis-like behavior. Although our results are largely empirical, the combination above yielded a control system that can be deployed directly on hardware without post-transfer fine-tuning and remains robust against disturbances in all tested scenarios.

\bibliography{references_ML_bibtex}

\newpage
\section{Appendix}

\begin{figure}[htpb]
    \centering
    \begin{subfigure}{0.25\textwidth}
        \includegraphics[width=\linewidth]{figures/SIP.jpg}
    \caption{Lab Photo}
    \label{fig:lab-sip}
    \end{subfigure}%
    \begin{subfigure}{0.75\textwidth}
        \centering
        \includegraphics[width=\linewidth]{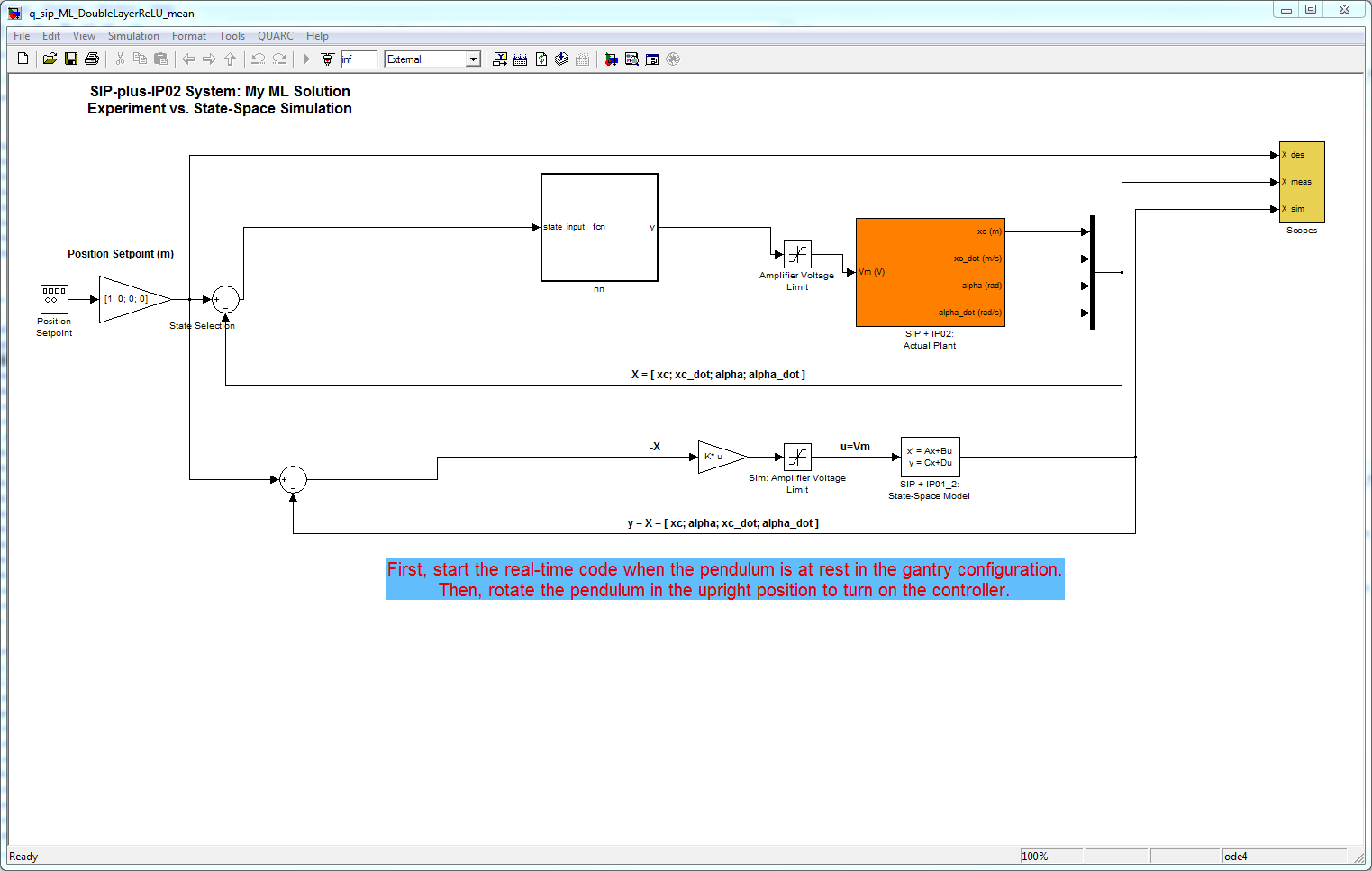}
        \caption{Simulink interface}
        \label{fig:stab-simulink-top}
    \end{subfigure}
    \caption{Inverted pendulum in the lab and the hardware-in-loop interface with Simulink}
\end{figure}

\begin{figure}
    \centering
    \begin{subfigure}[b]{0.49\textwidth}
        \includegraphics[width=\linewidth]{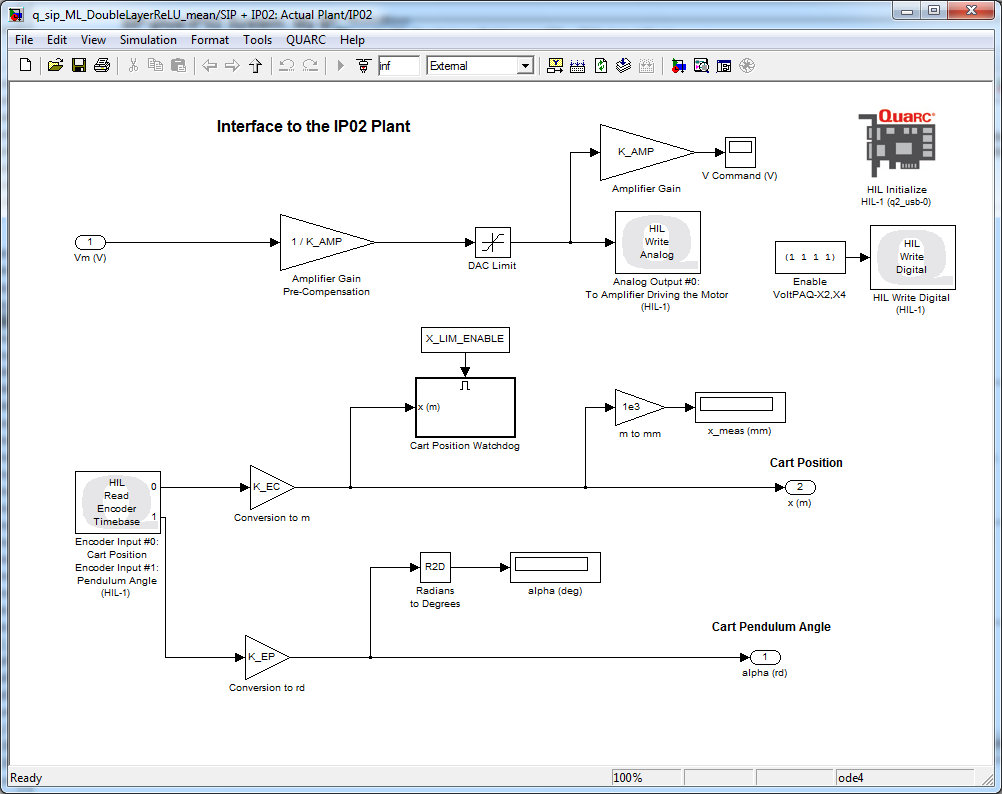}
    \caption{Simulink plant dynamics}
    \label{fig:simulink-plant-dynamics}
    \end{subfigure}%
    \begin{subfigure}[b]{0.49\textwidth}
        \centering
        \includegraphics[width=\linewidth]{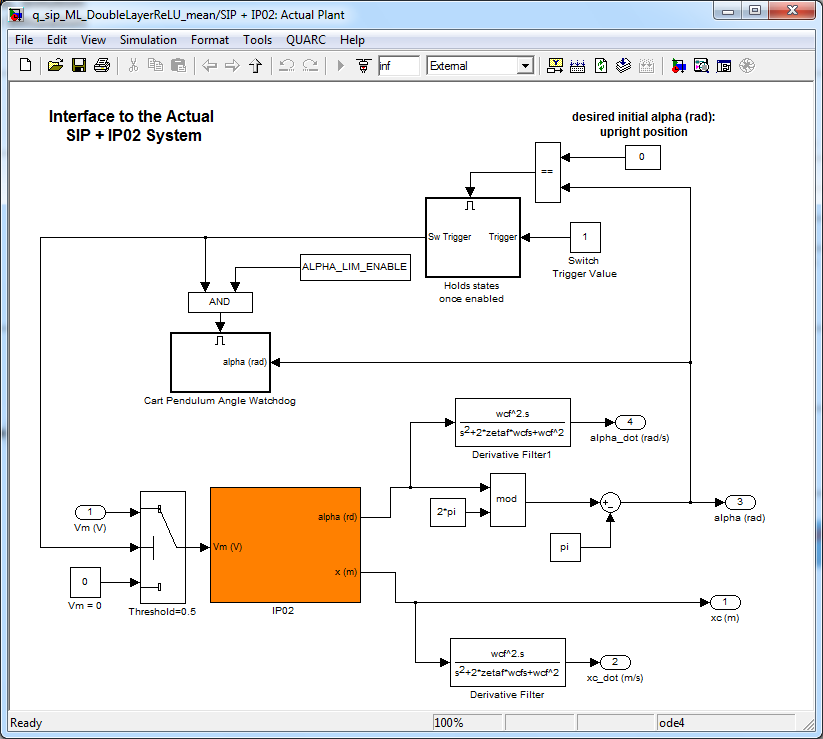}
        \caption{Simulink plant details}
        \label{fig:stab-simulink-plant}
    \end{subfigure}
    \caption{Simulink hardware-in-loop interfaces for controlling inverted pendulum}
\end{figure}

\end{document}